\documentclass[journal]{IEEEtran}
\usepackage{subcaption,booktabs}
\usepackage{siunitx} 
\usepackage{amsmath,amsfonts}
\usepackage{amssymb}
\usepackage{algorithmic}
\usepackage{algorithm}
\usepackage{array}
\usepackage{textcomp}
\usepackage{stfloats}
\usepackage{url}
\usepackage{verbatim}
\usepackage{graphicx}
\usepackage{cite}
\usepackage{multirow}
\usepackage{enumitem}
\usepackage{xcolor}
\usepackage{tabularx} 
\newtheorem{mythr}{Theorem}
\newtheorem{mylem}{Lemma}

\usepackage{hyperref}
\definecolor{custompink}{RGB}{239,118,186} 
\hypersetup{
    colorlinks=true, 
    linkcolor=blue,  
    citecolor=green, 
    filecolor=magenta, 
    urlcolor=custompink     
}

\title{QUADFormer: Learning-based Detection of Cyber Attacks in Quadrotor UAVs}

\author{Pengyu Wang$^{\dag, 1}$, \textit{Student Member, IEEE}, Zhaohua Yang$^{\dag, 2}$, Nachuan Yang$^{2}$, Zikai Wang$^{2}$, \\ Jialu Li$^{2}$, Fan Zhang$^{2}$, Chaoqun Wang$^{3}$, \textit{Member, IEEE}, Jiankun Wang$^{4}$, \textit{Senior Member, IEEE}, \\ Max Q.-H. Meng$^{5}$, \textit{Fellow, IEEE} and Ling Shi$^{2}$, \textit{Fellow, IEEE}

\thanks{$^{\dag}$ Equal contribution.}
\thanks{$^{1}$Pengyu Wang is with the Shenzhen Key Laboratory of Robotics Perception and Intelligence and the Department of Electronic and Electrical Engineering, Southern University of Science and Technology, Shenzhen, China. He is also with the Department of Electronic and Computer Engineering, Hong Kong University of Science and Technology, Hong Kong SAR. {\tt\small pwangat@connect.ust.hk}}
\thanks{$^{2}$Zhaohua Yang, Nachuan Yang, Zikai Wang, Jialu Li, Fan Zhang and Ling Shi are with the Department of Electronic and Computer Engineering, Hong Kong University of Science and Technology, Hong Kong SAR. {\tt\small \{zyangcr, nyangaj, zwanggz, jlikr, fzhangaw\}@connect.ust.hk, eesling@ust.hk}}
\thanks{$^{3}$Chaoqun Wang is with the School of Control Science and
Engineering, Shandong University, Shandong, China. {\tt\small chaoqunwang@sdu.edu.cn}}
\thanks{$^{4}$Jiankun Wang is with the Shenzhen Key Laboratory of Robotics Perception and Intelligence and the Department of Electronic and Electrical Engineering, Southern University of Science and Technology, Shenzhen, China. He is also with Jiaxing Research Institute, Southern University of Science and Technology, Jiaxing, China. {\tt\small wangjk@sustech.edu.cn}}
\thanks{$^{5}$Max Q.-H. Meng is with the Shenzhen Key Laboratory of Robotics Perception and Intelligence and the Department of Electronic and Electrical Engineering, Southern University of Science and Technology, Shenzhen, China. He is also a Professor Emeritus in the Department of Electronic Engineering at The Chinese University of Hong Kong in Hong Kong and was a Professor in the Department of Electrical and Computer Engineering at the University of Alberta in Canada. {\tt\small max.meng@ieee.org}}
}

\begin{document}

\maketitle
\thispagestyle{empty}
\pagestyle{empty}

\begin{abstract}
Safety-critical intelligent cyber-physical systems, such as quadrotor unmanned aerial vehicles (UAVs), are vulnerable to different types of cyber attacks, and the absence of timely and accurate attack detection can lead to severe consequences. When UAVs are engaged in large outdoor maneuvering flights, their system constitutes highly nonlinear dynamics that include non-Gaussian noises. Therefore, the commonly employed traditional statistics-based and emerging learning-based attack detection methods do not yield satisfactory results. In response to the above challenges, we propose QUADFormer, a novel Quadrotor UAV Attack Detection framework with transFormer-based architecture. This framework includes a residue generator designed to generate a residue sequence sensitive to anomalies. Subsequently, this sequence is fed into a transformer structure with disparity in correlation to specifically learn its statistical characteristics for the purpose of classification and attack detection. Finally, we design an alert module to ensure the safe execution of tasks by UAVs under attack conditions. We conduct extensive simulations and real-world experiments, and the results show that our method has achieved superior detection performance compared with many state-of-the-art methods. (Video\footnote{\href{https://youtu.be/bQSo0fjq54E}{https://youtu.be/bQSo0fjq54E}})

\end{abstract}


% \def\abstractname{Note to Practitioners}
% \begin{abstract}
% In the field of quadrotor UAVs, ensuring operational safety against cyber-attacks has paramount importance. Traditional detection methods often struggle in complex environments, frequently resulting in either missed attacks or false alarms. To address this, we have developed a new approach that significantly enhances the detection of stealthy and harmful cyber manipulations, such as those altering the flight data of UAVs. Our method begins by employing an extended Kalman filter to refine UAV sensor data, reducing noise and extracting meaningful statistical residues. Subsequently, these residues are analyzed using a transformer-based architecture, designed to capture temporal dependencies and anomalies indicative of cyber attacks. Our approach is highly effective in both simulated environments and real-world applications due to its ability to dynamically adapt to different flight conditions and attack types, consistently outperforming existing methods. For industry professionals, adopting this method means enhancing the security and reliability of UAV operations, protecting them from potential cyber threats, and ensuring that these vital tools can perform their tasks without interruption. This development represents a significant step forward in UAV technology, offering a novel solution to one of the most urgent challenges in UAV deployment.
% \end{abstract}

\section{Introduction}
\label{sec:introduction}

Cyber-physical system (CPS) is the integration of physical processes with cyber-infrastructure, where human operators monitor and control these processes using Supervisory Control and Data Acquisition (SCADA) systems~\cite{alur2015principles}, and it has become a critical component of modern society. Unmanned aerial vehicles (UAVs) are a typical example of intelligent CPS that can achieve different autonomous tasks by integrating smart sensors, communication networks, and flight control systems. They have been widely applied in practical scenarios, including environmental surveillance~\cite{li2019coverage}, search-and-rescue~\cite{ngo2022uav}, and air-ground collaboration~\cite{wang2022quadrotor}, to name just a few.

\begin{figure}[t!]
\centering
\includegraphics[width=0.98\columnwidth]{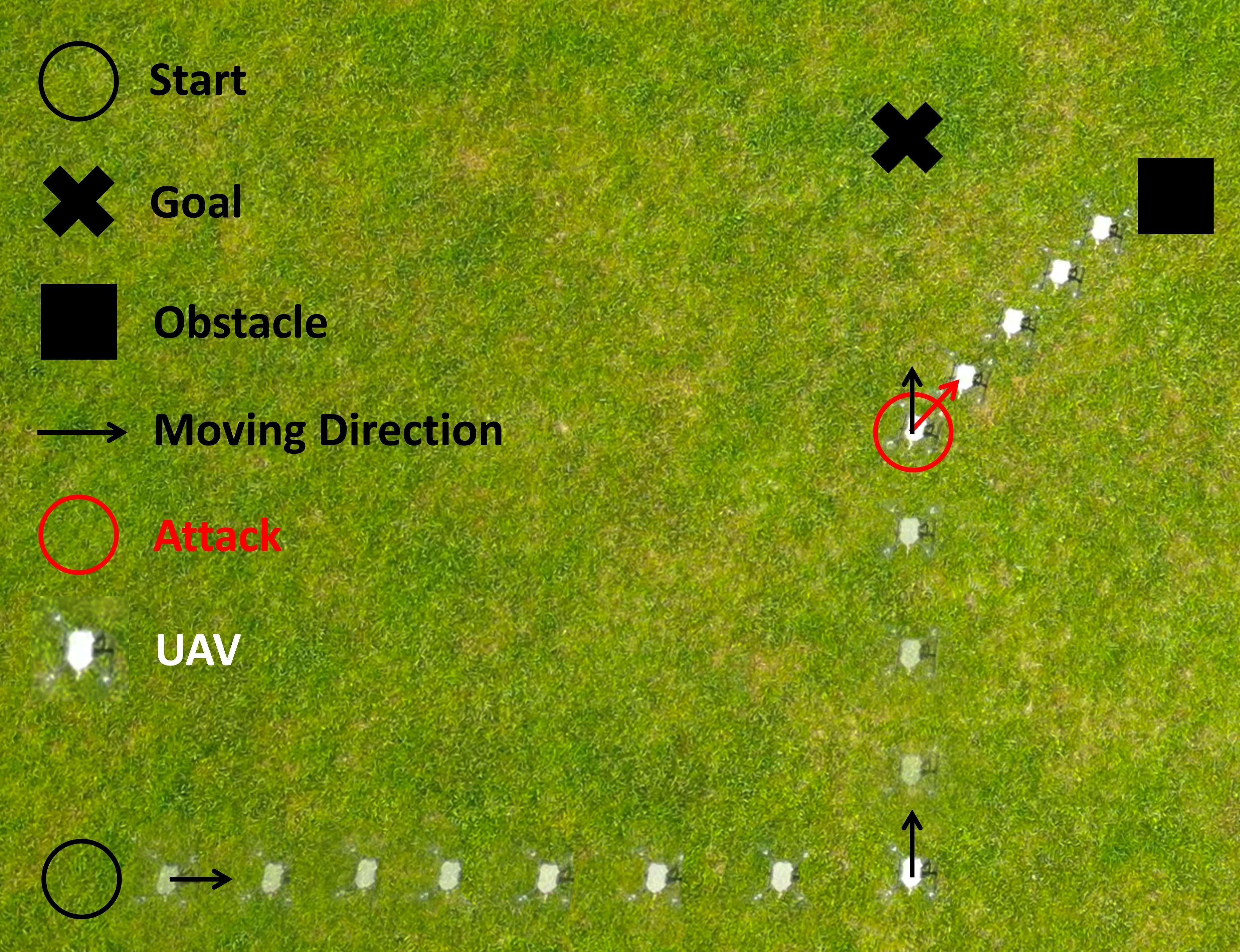} 
\caption{Illustration of UAV operation under cyber attacks.}
\label{fig:attack_overview}
\end{figure}

For safety-critical UAVs operating in real-world environments, their vulnerabilities could potentially lead to severe accidents. For example, the openness of communication networks, such as GPS, as shown in Fig.~\ref{fig:attack_overview}, makes UAVs particularly susceptible to cyber attacks~\cite{restivo2023gps}~\cite{adil2023uav}, and these challenges have already turned into reality, causing significant damages and even tragedies~\cite{shepard2012evaluation, kerns2014unmanned, hooper2016securing, dey2018security, gregory2013cyber, hassija2021fast}. In~\cite{shepard2012evaluation} and \cite{kerns2014unmanned}, the authors employ both overt and covert spoofing strategies to manipulate GPS measurements, successfully capturing and controlling the Hornet Mini UAV, leading to irrecoverable errors and its subsequent crash. For consumer-grade drones, the Parrot Bebop has been found to be susceptible to attacks during the ARDiscovery connection process~\cite{hooper2016securing}, and the DJI Phantom 4 has also been found to be vulnerable to jamming attacks through its SDK~\cite{dey2018security}. Moreover, military UAVs have been targeted with spoofing attacks on their control systems~\cite{gregory2013cyber}, among which the most notable is the capture of the U.S. Lockheed Martin RQ-170 Sentinel UAV by Iran through related attacks. 

Common cyber attacks in UAVs can be categorized into different types based on
their mechanisms, including denial-of-service (DoS) attacks, false data injection (FDI) attacks,
and replay attacks~\cite{griffioen2019tutorial}. Specifically, DoS attacks refer to the malicious disruption of information flows over communication networks~\cite{chen2019container}. FDI attacks involve corrupting the integrity of transmitted sensor or actuator data to affect the plant~\cite{chen2017false}. Replay
attacks correspond to hijacking the sensors for a certain amount of time and repeatedly replaying the same data. Detecting replay attacks requires continuous active methods such as watermark-based methods~\cite{porter2020detecting}~\cite{fang2020optimal}. FDI attacks, being more prevalent and easier to implement than replay attacks, also exhibit greater covert and deceptive characteristics compared to DoS attacks. 

To secure CPS against malicious attacks, researchers have designed various detection methods that can primarily be categorized into two types: residue-based detection and learning-based detection. Residue-based detection of FDI attacks in CPS uses statistical tests like Chi-squared and cumulative sum control chart (CUSUM)~\cite{murguia2016cusum}~\cite{murguia2016characterization}, analyzing filter residues to trigger alarms when thresholds are exceeded. Despite the effectiveness of different residue-based methods, they share two common and obvious limitations. First, their methods~\cite{yoon2019towards}~\cite{liu2019secure} rely on the low-dimensional data of the model or the assumption of Gaussian noise. The performance of the above methods significantly degrades when applied to UAVs with high-dimensionally nonlinear dynamics and non-Gaussian noises. This is because the statistical characteristics tend to become irregular, rendering related methods incapable of accurately capturing anomalies. Second, their methods~\cite{muniraj2017framework}~\cite{wan2020safety} are sensitive to dynamic changes. CUSUM-type methods need to adjust the optimal threshold according to specific data conditions. However, during the high maneuverability of UAVs, it is unrealistic to adjust an optimal threshold for every situation due to environmental changes and unknown attack types. This motivates us to adopt a learning-based approach to capture the high-order dynamic information and non-Gaussian noise properties in residue sequences. 

In recent years, researchers have also proposed several learning-based methods for attack detection. These methods~\cite{farivar2019artificial}~\cite{shen2020timeseries} leverage machine learning techniques to identify patterns and anomalies that may indicate the presence of an attack. However, previous learning-based methods primarily suffer from two issues. First, most of the approaches~\cite{shafique2021detecting}~\cite{khoei2022comparative} directly take raw sensor data as input, which encompasses different dimensions and arbitrary ranges and is susceptible to different noises, leading to low detection accuracy and high demand for large training datasets. Second, the majority of attack detection methods~\cite{wu2023highly}~\cite{wang2020intelligent} for sequential data only learn local point representations (e.g., SVM, CNN) or rely on recursive state transfer and do not have parallel capabilities (e.g., RNN, LSTM), resulting in limited contextual understanding and increased model complexity. This motivates us to adopt a more suitable input data source and, at the same time, use the transformer-based architecture that can effectively capture the dependencies of sequence data to achieve attack detection through a modified attention design. 

In order to address the aforementioned challenges, we combine the advantages of state estimation-based methods with those of learning-based approaches, proposing a novel attack detection framework called \textbf{QUADFormer}, which is specifically for \textbf{Q}uadrotor \textbf{U}AV \textbf{A}ttack \textbf{D}etection with a novel designed trans\textbf{Former} architecture. Unlike previous methods, we initiate our process by running an extended Kalman filter (EKF) on the UAV to collect residues (innovations) data. Residues, in contrast to raw data, contain less noise, more useful statistical information, and are of lower dimensionality. Subsequently, we perform feature engineering on the residue sequence data to facilitate the neural network's ability to better learn its statistical properties. Then, we use the residue data as input to design a transformer-based neural network architecture that learns the features within the residues and the discrepancy between anomalous and normal points for attack detection, where these features are highly dependent on the transition dynamics and noise characteristics. Finally, we design a resilient state estimation module based on the results of the attack detection to ensure that the UAV remains operational even in the presence of attacks. The main contributions of this paper are multifold:

\begin{enumerate}[label=\arabic*)]
  \item We propose a novel cyber attack detection framework for quadrotor UAVs, which combines the residue-based and learning-based methods and overcomes their existing shortcomings.
  \item We leverage the framework to effectively extract and learn the statistical characteristics of the UAV's highly nonlinear dynamics and non-Gaussian noises, thereby achieving outstanding sensor attack detection performance. 
  \item We conduct real, low-cost electronic attacks on UAVs and achieve resilient state estimation by deploying the proposed attack detection framework, compared to only simulated attacks in most previous research. The source code will be made publicly available.
\end{enumerate}

The remainder of the paper is organized as follows. We first review related work in Section \uppercase\expandafter{\romannumeral2}. In Section \uppercase\expandafter{\romannumeral3}, we give some preliminaries and present the problem formulation. Section \uppercase\expandafter{\romannumeral4} provides the details of our proposed algorithm. We conduct extensive comparative experiments and data analysis in Section \uppercase\expandafter{\romannumeral5}. Finally, we draw a conclusion and discuss our future work in Section \uppercase\expandafter{\romannumeral6}.

\section{Related Work}

\subsection{Residue-based Attack Detection}
In generally intelligent CPS, the detection of FDI attacks is typically approached from a statistical perspective. Chi-squared and CUSUM tests have been employed in previous studies~\cite{murguia2016cusum}~\cite{murguia2016characterization} to detect FDI attacks using residue information from Kalman filters. An alarm is triggered if the dissimilarity or the cumulative sum exceeds a predefined threshold. In the UAV system, Muniraj \textit{et al.}~\cite{muniraj2017framework} propose a statistics-based GPS spoofing attack detection framework for small UAVs, which includes implementations of the CUSUM, sequential probability ratio test (SPRT), and binary hypothesis test (BHT). Similarly, in~\cite{yoon2019towards} and \cite{wan2020safety}, the authors present modified CUSUM detectors with forgetting factors to achieve asymptotic behavior in GPS-denied environments. Recently, a modified sliding innovation sequences method has been proposed by Xiao \textit{et al.}~\cite{xiao2022cyber} to achieve detection and system isolation for various types of attacks. This method aims to address the issue of false alarms generated by the CUSUM approach during high maneuverability flights of UAVs, particularly at the moments immediately following the initiation of an attack. 

\subsection{Learning-based Attack Detection} 
In generally intelligent CPS, Farivar \textit{et al.}~\cite{farivar2019artificial} propose a framework combining nonlinear control and neural network for attack estimation and compensation. Furthermore, a statistical learning method has been developed in~\cite{yang2023lasso} to detect and identify actuator integrity attacks in remote control systems using the least absolute shrinkage and selection operator (LASSO). Researchers from the field of data science consider the relevant problem as anomaly/fault detection. In~\cite{shen2020timeseries} and~\cite{li2021multivariate}, researchers utilize neural networks to learn point representations in anomalous data sequences and determine the location of attacks through reconstruction errors. Tuli \textit{et al.}~\cite{tuli2022tranad} and Xu \textit{et al.}~\cite{xu2022anomaly} further propose a transformer-based~\cite{vaswani2017attention} detection network for multivariate time series data. In the UAV system, Shafique~\cite{shafique2021detecting} \textit{et al.} and Khoei~\cite{khoei2022comparative} \textit{et al.} both use support vector machine (SVM)-based supervised learning method for GPS spoofing detection. Abbaspour \textit{et al.}~\cite{abbaspour2017neural} present a neural adaptive approach to detect FDI attacks in both the sensor and actuator with stability guarantee. In~\cite{xiao2017user}, they propose a DQN-based power allocation strategy for a UAV designed to counteract different attacks without requiring prior knowledge of the attack model. Ma \textit{et al.}~\cite{ma2023detecting} create a UAV anomaly data set and use a method based on graphical normalizing flows for multivariate anomaly detection. Based on the previous work, Abbaspour \textit{et al.}~\cite{abbaspour2018neural} further design an active fault-tolerant control system with the application of FDI. In~\cite{hickling2023robust}, the authors propose a set of attacks and defenses for UAV systems using deep reinforcement learning for motion planning. Most recently, a generic framework for low-cost GPS spoofing attack detection has been proposed in~\cite{wu2023highly} and the authors use Shapley's additive explanations to increase the interpretability of different neural networks. 

\section{Preliminaries}

\subsection{UAV Model}
The general quadrotor UAV system is formulated as follows:
\begin{subequations}\label{eq:system_model}
    \begin{align}
    & x_{k+1} = f(x_k,u_k)+w_k \label{eq:state_update},\\
    & y_k=g(x_k)+v_k \label{eq:measurement_update},
    \end{align}
\end{subequations}
where \( x_k \in \mathbb{R}^n \) is the state vector, \( u_k \in \mathbb{R}^p \) is the control input vector, \( y_k \in \mathbb{R}^m \) is the measurement vector, all at time step \( k \), $f(\cdot)$ and $g(\cdot)$ are non-linear transition functions of the state model and measurement model, \( w_k \in \mathbb{R}^n \) and \( v_k \in \mathbb{R}^m \) are process noise and measurement noise vectors with covariance \( Q_k \in \mathbb{R}^{n \times n} \succeq 0 \) and \( R_k \in \mathbb{R}^{m \times m} \succ 0 \). 

\subsection{Extended Kalman Filter}
We present several preliminaries of the extended Kalman filter with respect to (\ref{eq:system_model}) when the system is not under cyber attack and the EKF process is shown as follows:
\begin{itemize}
    \item Predict: predicted state estimate and the covariance matrix.
    \begin{subequations}\label{eq:ekf_predict}
        \begin{align}
        & \hat x_{k|k-1} = f(\hat x_{k-1|k-1},u_{k-1}),\\
        \begin{split}
        & P_{k|k-1} = \nabla_x f(\hat x_{k-1|k-1},u_{k-1}) \cdot P_{k-1|k-1} \\
        & \quad\quad\quad \cdot \nabla_x f(\hat x_{k-1|k-1},u_{k-1})^T + Q_{k-1},
        \end{split}
        \end{align}
    \end{subequations}

    \item Update: updated state estimate and the covariance matrix.
    \begin{subequations}\label{eq:ekf_update}
        \begin{align}
        \begin{split}
        & K_k = P_{k|k-1} \cdot \nabla_xg(\hat x_{k|k-1})^T \\
        & \quad \cdot (\nabla_xg(\hat x_{k|k-1}) \cdot P_{k|k-1} \cdot \nabla_xg(\hat x_{k|k-1})^T + R_k)^{-1},
        \end{split} \\
        & \hat x_{k|k} = \hat x_{k|k-1} + K_k(y_k - g(\hat x_{k|k-1})), \\
        & P_{k|k} = P_{k|k-1} - K_k \cdot \nabla_xg(\hat x_{k|k-1}) \cdot P_{k|k-1},
        \end{align}
    \end{subequations}
\end{itemize}

The Jacobian matrices with respect to the state transition and measurement model, denoted by $\nabla_x f(\cdot) \in \mathbb{R}^{n \times n}$ and $\nabla_x g(\cdot) \in \mathbb{R}^{m \times n}$, serve the roles of state transition matrix and measurement matrix around the current estimate. The Kalman gain is \(K_k \in \mathbb{R}^{n \times m}\) and the residue term \(r_k = y_k - g(\hat x_{k|k-1}) \in \mathbb{R}^m\) reflects the difference between the actual measurements and the predicted measurements at time step \(k\).

\subsection{False Data Injection Attack}

In an FDI attack, the attacker manipulates the sensor readings, which could involve GPS spoofing or other forms of deception. The measurement equation under the FDI attack is:
\begin{equation}\label{eq:fdi}
    y_k^{i} = g^i(x_k)+ v_k^{i} + d_k^{i},
\end{equation}
where variables with the superscript \( i \) correspond to the \( i^{th} \) sensor and \( i \in \{1, 2, ..., m\} \). The vector \( d_k^{i} \in \mathbb{R}^m \) represents the attack vector, being a sparse and persistent false data injection on the \( i^{th} \) sensor.

Specifically, sparse attacks make occasional changes to sensor data in terms of full runtime, which can look like regular sensor errors, allowing them to slip past usual attack-detection systems. Persistent attacks keep altering sensor readings for a long time, slowly interfering with the UAV's operations and consistently feeding wrong information into the controller.

\subsection{Performance Metric and Problem Formulation}
To evaluate the performance of our attack detector, we require some performance metrics. On the one hand, when the attack is vital to the system, we hope the detector can promptly trigger an alarm and our primary focus is on achieving high sensitivity. On the other hand, in most situations, we do not want the system to stop functioning every time an alarm is triggered, especially when the robotic system is designed for long-term operation. Typically, the system is equipped with multiple redundant sensors, and when an alarm occurs for one sensor, that specific sensor is temporarily deactivated. The typical performance indices are defined as follows. 

\textbf{Long-term Attack Detection}\hspace{5pt} The main criteria used are as follows:
\begin{equation}\label{eq:precision}  
    \textit{Precision} = \frac{\text{TP}}{\text{TP} + \text{FP}},  
\end{equation}
\begin{equation}\label{eq:recall}  
    \textit{Recall} = \frac{\text{TP}}{\text{TP} + \text{FN}},  
\end{equation}  
\begin{equation}\label{eq:F1}
    \textit{F1-score} = \frac{2}{\textit{Precision}^{-1} + \textit{Recall}^{-1}},
\end{equation}
where in our cases, \textit{Precision} (P) is the ratio of the number of correctly alarmed time steps to the total number of alarmed time steps, \textit{Recall} (R) is the ratio of the number of correctly alarmed time steps to the total number of attacked time steps, and \textit{F1-score} (F1) is the harmonic mean of them.

\textbf{Problem Formulation}\hspace{5pt} Consider system (\ref{eq:system_model}) with possible attacks (\ref{eq:fdi}) acting on the system, design an attack detector such that the performance metrics in (\ref{eq:precision}) - (\ref{eq:F1}) are maximized.  

\section{Methodology}

\begin{figure*}[ht!]
\centering
\includegraphics[width=\textwidth]{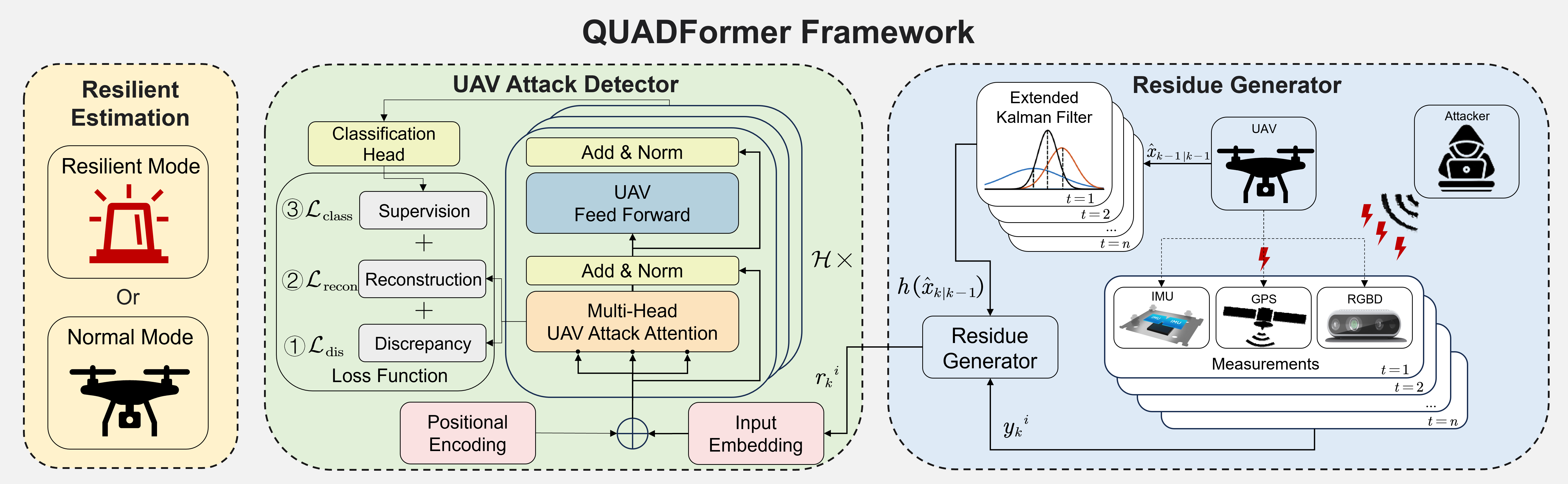}
\caption{Overview of QUADFormer framework. The framework starts with an effective residue generator (right, Sec.~\ref{sec:residue_generator}). The residue sequences are then fed into a novel attack detector (middle, Sec.~\ref{sec:attack_detector}). The UAV achieves resilient state estimation based on the detection results (left, Sec.~\ref{sec:state_estimation}).}
\label{fig:frame}
\end{figure*}

Our QUADFormer framework mainly includes a residue generator, a UAV attack detector and a resilient state estimation module, as shown in Fig.~\ref{fig:frame}. First, we employ the EKF to fuse multisensor data, which allows the UAV to estimate the system's state and generate residue sequences containing a large amount of pattern information. Second, we devise a novel semi-supervised transformer-based architecture with a specially designed self-attention module and loss functions, which is used for classification and subsequent detection of malicious attacks. Finally, we design a resilient state estimation module to promptly shut down the sensors under attack and use secure sensors to fuse and estimate the pose. The module then promptly reactivates the sensors when they are secure again.

\subsection{UAV Residue Generator}\label{sec:residue_generator}  

In the research field of state estimation, there are mainly two ways to generate residue information (difference between observed values and model predictions): filter-based methods and optimization-based methods. When these two methods are under the framework of Bayesian probability and the Gaussian-Newton algorithm, the mathematical logic of the two methods is essentially equivalent. In this paper, since the state dimension of the system is not large and the filter-based method has better numerical stability, we choose to use a filter-based method to generate residue sequences for the UAV. Compared with other nonlinear filtering methods, EKF has lower computational complexity, as it only requires calculating the Jacobian matrix. Unlike the Unscented Kalman Filter (UKF), it does not need to capture the probability distribution of the state through Sigma points~\cite{menegaz2015systematization}, nor does it use a large number of particles to approximate the posterior distribution like the Particle Filter~\cite{gustafsson2010particle}. In addition to the advantages introduced in Sec.~\ref{sec:introduction} compared with the raw sensor data, the residue data generated through the EKF update process also has two most important characteristics, making EKF residue highly suitable for UAV attack detection. First, the calculation of model predictions introduces first-order linearization through Taylor expansion and ignores higher-order terms, so that the overall residue term contains high-order information, and the high-order information reflects the dynamics characteristics of nonlinear UAV systems. Second, EKF is directly derived from KF and has a more intuitive mathematical expression. At the same time, the operation of EKF is based on the assumption of Gaussian noise. Therefore, when the actual noise is single-mode non-Gaussian noise (multi-modal non-Gaussian noise is very rare in real situations), the resulting residue sequence has near-Gaussian properties and contains an approximation error to the Gaussian noise itself. Based on the above points, EKF's residue information has good statistical properties of UAV nonlinear systems and non-Gaussian noise, making it suitable for feature learning by subsequent neural networks. In what follows, we present some results on the boundedness of EKF's residues, and we first define the following functions:
\begin{align*}
    &\varphi \left( x_k,\hat{x}_{k|k},u_k \right):= f(x_k,u_k) - f(\hat{x}_{k|k},u_k) \\
    &~~~~~~~~~~~~~~~~~~~~~~~~~~~~~~~~~~- \nabla_x f(\hat{x}_{k|k},u_k)(x_k-\hat{x}_{k|k})\\
    &\chi \left( x_k,\hat{x}_{k|k} \right):= g(x_k) - g(\hat{x}_{k|k}) - \nabla_x g(\hat{x}_{k|k}) (x_k-\hat{x}_{k|k})
\end{align*}
which represent the high-order terms of the Taylor expansions of $f(\cdot)$ and $g(\cdot)$ at the point $\hat{x}_{k|k}$, respectively.
\vspace{3pt}

\begin{mylem}\cite{reif1999stochastic}
Consider the UAV system given by (\ref{eq:system_model}) and the extended Kalman filter described in (\ref{eq:ekf_predict}) - (\ref{eq:ekf_update}). If

\begin{enumerate}[label=\arabic*)]
  \item There exist $\beta _1, \beta _2, \beta _3, \beta _4 > 0$ such that
  \begin{subequations}
    \begin{align}
    & \left\| \nabla _xf(\hat{x}_{k|k}) \right\| \leqslant \beta _1,\\
    & \left\| \nabla _xg(\hat{x}_{k|k}) \right\| \leqslant \beta _2, \\
    & \beta _3\leqslant P_{k|k}\leqslant \beta _4,\forall k\ge 0.
    \end{align}
  \end{subequations}
  \item $\nabla_x f(\cdot)$ is non-singular for all $k\ge 0$.
  \item There exist $\epsilon _{\varphi}, \epsilon _{\chi}, \iota _{\varphi}, \iota _{\chi}>0$ such that 
  \begin{subequations}
    \begin{align}
    \left\| \varphi \left( x_k,\hat{x}_{k|k},u_k \right) \right\| &\le \epsilon _{\varphi}\left\| x_k-\hat{x}_{k|k} \right\| ^2,\\
    \left\| \chi \left( x_k,\hat{x}_{k|k} \right) \right\| &\le \epsilon _{\chi}\left\| x_k-\hat{x}_{k|k} \right\| ^2,
    \end{align}
  \end{subequations}
  where $\left\| x_k-\hat{x}_{k|k} \right\| \le \iota _{\varphi}$ and $\left\| x_k-\hat{x}_{k|k} \right\| \le \iota _{\chi}$.
\end{enumerate}
Then the estimation error is bounded in mean square, and moreover it is almost sure bounded, i.e.,
\begin{align}
   \exists M>0, \mathbb{P}(\|x_k-\hat{x}_{k|k}\|<M) = 1,
\end{align}
\end{mylem}
provided that the initial estimation error and the covariance matrices of the noises are bounded.

Proof. The proof follows from Theorem 3.1 in \cite{reif1999stochastic} with a few modifications based on the UAV setup in this paper. $\hfill \square$
\vspace{3pt}

\begin{mythr}
    Given that $f(\cdot),g(\cdot)$ are Lipschitz continuous and the conditions 1) to 3) in \textit{Lemma} 1 hold, the EKF's residue $r_k := y_k - g(\hat x_{k|k-1})$ is almost sure bounded, i.e.,
    \begin{align}
    \exists N>0,\mathbb{P}(\|r_k\|<N) = 1.
\end{align}
\end{mythr}
Proof. Since $g(\cdot)$ is Lipschitz continuous, it holds that 
\begin{align*}
    \|y_k - g(\hat x_{k|k-1})\| &=\|g(x_k)+v_k- g(\hat x_{k|k-1})\| \\
    &=\|g(x_k)- g(\hat x_{k|k-1})\|+\|v_k\| \\
    &\le L_g\|x_k-\hat x_{k|k-1}\| + \|v_k\|
\end{align*}
where $L_g$ is the Lipschitz parameter of function $g(\cdot)$. Meanwhile, since $f(\cdot)$ is Lipschitz continuous, it holds that
\begin{align*}
   &\|x_k-\hat x_{k|k-1}\|\\
   &\le \|f(x_{k-1},u_{k-1})+w_{k-1} -f(\hat{x}_{k-1|k-1},u_{k-1})\|\\
    &\le \|f(x_{k-1},u_{k-1})-f(\hat{x}_{k-1|k-1},u_{k-1})\| +\|w_{k-1}\|\\
    &\le L_f\|x_{k-1}-\hat{x}_{k-1|k-1}\| +\|w_{k-1}\|
\end{align*}
where $L_f$ is the Lipschitz parameter of function $f(\cdot)$. Therefore we have $\|r_k\| = \|y_k - g(\hat x_{k|k-1})\| \le L_gL_f\|x_{k-1}-\hat{x}_{k-1|k-1}\|+L_g\|w_{k-1}\|+\|v_k\|.$
Notice that $\|x_{k-1}-\hat{x}_{k-1|k-1}\|$ is almost sure bounded according to Lemma 1 and the noises $w_k,v_k$ are almost sure bounded by physics, thus we can conclude that the residue $r_k$ is almost sure bounded. $\hfill \hfill \square$

\vspace{3pt}

\begin{algorithm}
\renewcommand{\algorithmicrequire}{\textbf{Input:}}
\renewcommand{\algorithmicensure}{\textbf{Output:}}
\caption{QUADFormer: UAV Residue Generator}
\label{alg:alg1}
\begin{algorithmic}[1] % 数字[1]表示行号从1开始
\REQUIRE Sensor data $\mathcal{S}$: GPS, Camera, IMU
\ENSURE Residue sequence $\mathcal{R} \in \mathbb{R}^{l \times m}$
\STATE $\mathcal{S} \gets$\textbf{TimeStampAlignment}($\mathcal{S}$)
\STATE $\mathcal{S} \gets$\textbf{CoordinateTransformation}($\mathcal{S}$)
\STATE \textbf{Initialization}($\hat{x}_{0|0}$, $P_{0|0}$, $\mathcal{R} \gets \left\{  \right\}$)
\FOR{$k=1,2,\ldots, l$}
    \STATE $\hat x_{k|k-1}$, $P_{k|k-1} \gets$ \textbf{Predict}($\hat x_{k-1|k-1}, u_{k-1}, P_{k-1|k-1}$)
    \STATE $K_k, \hat x_{k|k}, P_{k|k}, r_k \gets$ \textbf{Update}($\hat x_{k|k-1}, P_{k|k-1}, \mathcal{S}$)
    \STATE $\mathcal{R} \gets \mathcal{R} \cup \{r_k\}$ 
\ENDFOR
\STATE Return $\mathcal{R}$
\end{algorithmic}
\end{algorithm}

The above theoretical results characterize the stochastic stability of the residues obtained from the EKF. It is worth noting in {\it Theorem 1} that the residue is bounded from above by a weighted sum of the EKF's estimator error and process noises that are all close to zero, and the weights depend on the Lipschitzness of $g(\cdot)$ and $h(\cdot)$, which are independent of the actual values of $u_k, x_k, y_k$. Therefore, using residue data for attack detection is more appropriate than directly using sensor and actuator data, as done in many existing works. We will show in Sec.~\ref{subsec:system_modeling} that the two common UAV models that we use satisfy the three conditions in \textit{Lemma 1}. The whole procedure is shown in Alg.~\ref{alg:alg1} and the right part of Fig.~\ref{fig:frame}. For the first UAV model (\ref{eq:model1state}) - (\ref{eq:model1g}), the residue term $r_k \in \mathbb{R} ^{6\times 1}$ is calculated as:
\begin{equation}
r_k=\left[ \begin{array}{l}
	\mathbf{p}_k\\
	\mathbf{q}_k\\
\end{array} \right]-\left[ \begin{array}{l}
	\hat{\mathbf{p}}_k\\
	\hat{\mathbf{q}}_k\\
\end{array} \right] +v_k,
\end{equation}
where $\mathbf{p}_k\in \mathbb{R} ^{3\times 1}$ and $\mathbf{q}_k\in \mathbb{R} ^{3\times 1}$ are the newly observed position and orientation, $\hat{\mathbf{p}}_k\in \mathbb{R} ^{3\times 1}$ and $\hat{\mathbf{q}}_k\in \mathbb{R} ^{3\times 1}$ are the predicted position and orientation, $v_k \in \mathbb{R} ^{6\times 1}$ is the measurement noise. For the second UAV model (\ref{eq:model2state}) - (\ref{eq:model2g}), the residue term $r_k \in \mathbb{R} ^{6\times 1}$ is calculated as:
\begin{equation}
r_k=\left[ \begin{array}{l}
	\mathbf{p}_k\\
	\mathbf{q}_k\\
\end{array} \right] -\left[ \begin{array}{c}
	\mathbf{R}\left( \mathbf{q}_{\mathrm{F}} \right) ^T\left( \hat{\mathbf{p}}_k-\mathbf{p}_{\mathrm{F}} \right)\\
	\mathrm{Euler}\left( \mathbf{R}\left( \mathbf{q}_{\mathrm{F}} \right) ^T\mathbf{R}(\hat{\mathbf{q}}_k) \right)\\
\end{array} \right] +v_k,
\end{equation}
where $\mathbf{p}_{\mathrm{F}} \in \mathbb{R} ^{3\times 1}$ is the keyframe position, $\mathbf{q}_{\mathrm{F}} \in \mathbb{R} ^{3\times 1}$ is the keyframe orientation, $\mathrm{Euler}(): \mathbb{R} ^{3\times 3}\mapsto \mathbb{R} ^{3\times 1}$ converts rotation matrix into Euler angles.

\subsection{UAV Attack Detector}\label{sec:attack_detector}  

\subsubsection{Overall Architecture}

Our approach leverages the strengths of the transformer structure for processing sequential data and introduces the QUADFormer architecture, a dedicated framework tailored for quadrotor UAV attack detection. This architecture is developed based on some other transformer~\cite{tuli2022tranad}~\cite{xu2022anomaly} frameworks designed for anomaly detection, taking into account the specific characteristics of UAVs and their data. Specifically, we first create an attention module to more effectively capture the statistical characteristics of the attacked points within the UAV residue sequence data. Second, we develop a novel loss function that enables efficient semi-supervised learning. Finally, we establish new criteria for determining the occurrence of an attack. The overall diagram is shown in the middle part in Fig.~\ref{fig:frame} and the whole algorithm is shown in Alg.~\ref{alg:alg2}.

\begin{algorithm}
\renewcommand{\algorithmicrequire}{\textbf{Input:}}
\renewcommand{\algorithmicensure}{\textbf{Output:}}
\caption{QUADFormer: UAV Attack Detector}
\label{alg:alg2}
\begin{algorithmic}[1]
\REQUIRE Residue sequence $\mathcal{R} \in \mathbb{R}^{l \times m}$
\ENSURE Attack detection sequence $\mathcal{A} \in \mathbb{R}^{l \times 1}$ 

\STATE \textbf{Initialization}($\mathcal{Q}, \mathcal{K}, \mathcal{V}, \gamma, p$) 

\FOR{$h = 1$ \TO $\mathcal{H}$}
    \STATE $\mathcal{Q}^{h} \gets \mathcal{R}^h W^Q_h$, $\mathcal{K}^{h} \gets \mathcal{R}^h W^K_h$, $\mathcal{V}^{h} \gets \mathcal{R}^h W^V_h$
    \FOR{$i = 1$ \TO $l$}
        \FOR{$j = 1$ \TO $l$}
            \STATE $D_{i,j} \gets |j - i|^p$
            \STATE $\mathcal{T}_{i,j}^h \gets \exp(-\gamma \cdot D_{i,j})$ 
        \ENDFOR
        \STATE $\mathcal{T}_{i,:}^h \gets \frac{\mathcal{T}_{i,:}^h}{\sum_{k=1}^l \mathcal{T}_{i,k}^h}$
    \ENDFOR
    \IF{$h = 1$}
        \FOR{$i = 1$ \TO $l$}
            \STATE $\mathcal{C}_{i,:}^{h} \gets \mathrm{Softmax}\left( \frac{\mathcal{Q}_{i,:}^{h} \mathcal{K}^{h\mathrm{T}}}{\sqrt{d_{\mathrm{model}}}} \right)$ 
        \ENDFOR
    \ELSE
        \FOR{$i = 1$ \TO $l$}
            \STATE $\mathcal{C}_{i,:}^{h} \gets \mathrm{Softmax}\left( \frac{\mathcal{Q}_{i,:}^{h} (\mathcal{K}^{h-1})^{\mathrm{T}}}{\sqrt{d_{\mathrm{model}}}} \right)$ 
        \ENDFOR
    \ENDIF
\ENDFOR

\STATE $\mathrm{TCD(}\mathcal{T} ,\mathcal{C} ;\mathcal{R} ) \gets 0$ 
\FOR{$i = 1$ \TO $l$}
    \FOR{$h = 1$ \TO $\mathcal{H}$}
        \STATE $\mathrm{TCD}_i \gets \mathrm{TCD}_i + \mathrm{JS}\left(\mathcal{T}^h_{i,:}, \mathcal{C}^h_{i,:}\right)$
    \ENDFOR
    \STATE $\mathrm{TCD}_i \gets \frac{\mathrm{TCD}_i}{\mathcal{H}}$ 
\ENDFOR

\STATE $\mathcal{A} \gets \mathbf{0}_{l \times 1}$ 
\STATE Score $\mathcal{A}^s \in \mathbb{R}^{l \times 1} \gets \mathrm{Softmax}(-\mathrm{TCD}, \mathrm{dim}=0)$ 
\FOR{$i = 1$ \TO $l$}
    \IF{$\mathcal{A}^s_i > \text{threshold}$}
        \STATE $\mathcal{A}_i \gets 1$ 
    \ENDIF
\ENDFOR
\STATE Return $\mathcal{A}$ 

\end{algorithmic}
\end{algorithm}

\subsubsection{UAV Attack Attention}

We introduce a distinctive component of our QUADFormer architecture aimed at enhancing attack detection on quadrotor UAVs. This module is specially designed to interpret the intricacies of UAV dynamics by analyzing the residues derived from the previous EKF. We devise two novel constructs: Temporal Proximity Concentration (TPC) and Dynamic Context Mapping (DCM), which provide a refined approach to understanding UAV operational patterns.

\textbf{Initialization}\hspace{5pt} 
Our model processes an input sequence of residues $\mathcal{R} =\left\{ r_1,r_2,\ldots ,r_l \right\} \in \mathbb{R}^{l \times m}$, with each residue $r_k = y_k - g(\hat x_{k|k-1}) \in \mathbb{R}^m$. This sequence, spanning $l$ time steps, reflects the temporal dynamics of the UAV's operational state. The architecture consists of $\mathcal{H}$ layers, and thus, the input at each layer $h\in \left\{ 1,\ldots ,\mathcal{H} \right\} $ is represented as $\mathcal{R}^h \in \mathbb{R}^{l \times d_{\mathrm{model}}}$. Then, the self-attention is initialized with query ($\mathcal{Q}$), key ($\mathcal{K}$), and value ($\mathcal{V}$) matrices through transformations with learnable parameters. For each layer $h$, these matrices are sized $\mathcal{Q}^{h}, \mathcal{K}^{h}, \mathcal{V}^{h} \in \mathbb{R}^{l \times d_{\mathrm{model}}}$. These are obtained as $\mathcal{Q}^{h} = \mathcal{R}^h W^Q_h$, $\mathcal{K}^{h} = \mathcal{R}^h W^K_h$, and $\mathcal{V}^{h} = \mathcal{R}^h W^V_h$, where $W^Q_h, W^K_h, W^V_h \in \mathbb{R}^{d_{\mathrm{model}} \times d_{\mathrm{model}}}$ are the corresponding weight matrices for each layer $h$. Besides, the kernel width $\gamma \in \mathbb{R}^{l \times 1}$ is also initialized with respect to the input sequence $\mathcal{R}^h$. The size of the input and output for each transformer layer is determined by the hyperparameter $d_{\mathrm{model}}$.

\textbf{Temporal Proximity Concentration}\hspace{5pt} 
Observations within UAV flight data exhibit a natural continuity where adjacent temporal readings are more likely to share similar characteristics, especially in the event of an attack. To capitalize on this aspect, we introduce TPC, which is formally defined as:

\begin{equation}\label{eq:TPC}
    \mathcal{T} ^h=\mathrm{Rescale}\left( \left[ \exp \left( -\gamma \cdot |j-i|^p \right) \right] _{i,j\in \{1,\ldots ,l\}} \right),
\end{equation}
where $\gamma$ serves as a learnable scaling factor, and $p$ adjusts the focus on adjacent residues, with larger values narrowing the attention to immediate neighbors. $\mathcal{T}^h \in \mathbb{R}^{l \times l}$, corresponding to the pairwise temporal attention across all time steps in the sequence. $i$ and $j$ denote the indices of time steps within the sequence, with $i$ representing the current time step and $j$ representing another time step in the sequence. The Rescale operation ensures that the attention weights across different time steps are normalized. TPC thereby encodes the expectation that anomalies are temporally clustered, ensuring the model's heightened sensitivity to potential attacks.

\textbf{Dynamic Context Mapping}\hspace{5pt}
DCM is a technique enhanced to unravel the complex temporal patterns present in UAV flight data by mapping the interactions between time points across the sequence. In addition to self-attention scores, DCM integrates cross-layer attention similar to~\cite {chen2021crossvit}, allowing each layer to access information from the previous layer's key representations. For the first layer ($h=1$), since there is no previous layer to draw from, we use its own key matrix:
\begin{equation}\label{eq:first_layer_attention}
\mathcal{C}^1 = \mathrm{Softmax}\left( \frac{\mathcal{Q}^1 (\mathcal{K}^1)^{\mathrm{T}}}{\sqrt{d_{\mathrm{model}}}} \right).
\end{equation}

Then, the cross-layer attention for layer $h$ is generated as follows, assuming $h > 1$:
\begin{equation}\label{eq:cross_layer_attention}
\mathcal{C}^h = \mathrm{Softmax}\left( \frac{\mathcal{Q}^h (\mathcal{K}^{h-1})^{\mathrm{T}}}{\sqrt{d_{\mathrm{model}}}} \right),
\end{equation}
where $\mathcal{C}^h \in \mathbb{R}^{l \times l}$ captures the cross-layer attention for layer $h$, $\mathcal{Q}^h$ is the query matrix for the current layer, and $\mathcal{K}^{h-1}$ is the key matrix from the previous layer. This cross-layer strategy highlights patterns and dependencies that evolve over the sequence and across different layers of the network. The DCM module is expected to benefit from a richer representation of the UAV's flight data, as it can access a broader context from the preceding layers' features. This is particularly useful for capturing long-range dependencies and complex temporal dynamics in the flight data.

Through TPC and DCM, our UAV Attack Attention module enhances the transformer's ability to detect attacks by providing a more focused and contextually aware analysis of the UAV's residue sequences. Subsequently, the computed results are passed to layer $h+1$ for repeating the aforementioned calculations, which can be simply expressed by the following equations:
\begin{align}
\mathcal{Z}^{h+1} &= \mathrm{Norm}\left( \mathrm{UAV\ Attack\ Attention}\left( \mathcal{R}^{h} \right) + \mathcal{R}^{h} \right), \\
\mathcal{R}^{h+1} &= \mathrm{Norm}\left( \mathrm{Feed\ Forward}\left( \mathcal{Z}^{h+1} \right) + \mathcal{Z}^{h+1} \right),
\end{align}
where $\mathcal{Z}^h$ is the hidden representation with dimensions $\mathbb{R}^{l \times d_{\mathrm{model}}}$. The output of the final attention layer is denoted as $\widehat{\mathcal{R} }\in \mathbb{R} ^{l\times m}=\mathcal{R} ^{\mathcal{H}}=\mathcal{C} ^{\mathcal{H}}\mathcal{V} ^{\mathcal{H}}$, which is the reconstruction of input sequence $\mathcal{R}$.

\subsubsection{Loss Function}

We need a metric that better reflects the difference between the above two distributions, and then we define our novel loss function based on this metric.

\textbf{Temporal-Contextual Disparity}\hspace{5pt}
Temporal-Contextual Disparity (TCD) is a metric that quantifies the divergence between local and global temporal patterns using the Jensen-Shannon (JS) divergence. It is computed as:
\begin{equation}
    \mathrm{TCD}(\mathcal{T} ,\mathcal{C} ;\mathcal{R} )=\left[ \frac{1}{H}\sum_{h=1}^H{J}S\left( \mathcal{T} _{i,:}^{h}\parallel \mathcal{C} _{i,:}^{h} \right) \right] _{i=1,\ldots ,l},
\end{equation}
where $\mathcal{T} _{i,:}^{h}$ and $\mathcal{C} _{i,:}^{h}$ represent the TPC and DCM distributions for time step $i$ and layer $h$. Averaging over $H$ layers, TCD is always bounded and thus more stable in practice compared to the Kullback-Leibler (KL) divergence.

\textbf{Loss Function}\hspace{5pt}
Our loss function consists of two parts: one is the unsupervised error based on reconstruction, and the other is the supervised signal based on a small amount of labeled attack data. First, based on TCD, the discrepancy loss function is expressed as follows:
\begin{equation}
\mathcal{L}_{\mathrm{dis}}(\mathcal{T}, \mathcal{C}; \mathcal{R}) = \lambda \times \left\| \mathrm{TCD}(\mathcal{T}, \mathcal{C}; \mathcal{R}) \right\|_1,
\end{equation}
where $\parallel \cdot \parallel _1$ denotes $L_1$ norm and $\lambda$ is used for different training phase. Second, the first output head of the transformer represents the unsupervised reconstruction loss, which is expressed as follows:
\begin{equation}
\mathcal{L}_{\mathrm{recon}}(\widehat{\mathcal{R}}; \mathcal{R}) = \left\| \widehat{\mathcal{R}} - \mathcal{R} \right\|_{\mathrm{F}}^{2},
\end{equation}
where $\parallel \cdot \parallel _{\mathrm{F}}$ denotes Frobenius norm. Subsequently, to leverage the limited supervised label information, we introduce a classification loss at the second output head of the transformer:
\begin{equation}
\mathcal{L}_{\text{class}}(\widehat{\mathcal{Y}}, \mathcal{Y}) =  \sum_{i=1}^{N} -\mathcal{Y}_i \log(\widehat{\mathcal{Y}}_i) - (1-\mathcal{Y}_i) \log(1-\widehat{\mathcal{Y}}_i),
\end{equation}
where $\widehat{\mathcal{Y}}$ represents the model's prediction of anomalies, obtained by passing $\widehat{\mathcal{R}}$ through a classification head with a Sigmoid function. $\mathcal{Y}$ is the true label information, and $N$ is the number of labeled samples. Finally, similar to~\cite{xu2022anomaly}, we introduce a MinMax training strategy for $\mathcal{L}_{\mathrm{dis}}$. For the minimization phase, the total loss function to be minimized is represented as:
\begin{equation}
\mathcal{L}_{\mathrm{Total}} = \mathcal{L}_{\mathrm{recon}} + \mathcal{L}_{\mathrm{dis}} + \alpha_1 \mathcal{L}_{\mathrm{class}},
\end{equation}
where $\alpha_1$ is the balancing coefficient. For the maximization phase, the total loss function to be minimized is represented as:
\begin{equation}
\mathcal{L}_{\mathrm{Total}} = \mathcal{L}_{\mathrm{recon}} - \mathcal{L}_{\mathrm{dis}} + \alpha_2 \mathcal{L}_{\mathrm{class}},
\end{equation}
where $\alpha_2$ is also the balancing coefficient. Such a design allows the model to perform unsupervised reconstruction while also utilizing limited annotated data for supervised learning, thereby enhancing the detection performance of UAV systems.

\subsection{UAV Resilient State Estimation}\label{sec:state_estimation}

\begin{algorithm}
\renewcommand{\algorithmicrequire}{\textbf{Input:}}
\renewcommand{\algorithmicensure}{\textbf{Output:}}
\caption{QUADFormer: UAV Resilient State Estimation}
\label{alg:alg3}
\begin{algorithmic}[1] % 数字[1]表示行号从1开始
\REQUIRE Attack detection sequence $\mathcal{A} \in \mathbb{R}^{l \times 1}$
\ENSURE Adjusted pose estimation $\mathcal{X} \in \mathbb{R}^{l \times n}$
\FOR{$i = 1$ \TO $l-1$}
    \STATE $\mathcal{X}_{i+1} \gets \text{Fuse}(\text{GPS}, \text{camera}, \text{IMU})$ 
    \IF{$\mathcal{A}_i = 1$}
        \STATE $\mathcal{X}_{i+1} \gets \text{Fuse}(\text{camera}, \text{IMU})$ 
    \ENDIF
\ENDFOR
\STATE Return $\mathcal{X}$
\end{algorithmic}
\end{algorithm}

We integrate a resilient state estimation system into our UAV to swiftly counteract sensor attacks, as shown in Alg.~\ref{alg:alg3}. This system strikes a balance between constant vigilance and efficient response. Without any alarms, the system fuses data from GPS, camera, and IMU sensors for pose estimation:
\begin{equation}
\mathcal{X} = \text{Fuse}(\text{GPS}, \text{camera}, \text{IMU}),  
\end{equation}
where $\mathcal{X} \in \mathbb{R}^{l \times n}$ is the UAV pose of the currently considered sequence. Although visual-inertial odometry achieves good results, drift is inevitable when flying long distances, so incorporating readily available GPS absolute positioning information is a good correction. UAVs equipped with a GPS receiver, a stereo camera and an IMU are the most common and cost-effective solution for flying outdoors. The openness of GPS signal communication makes UAVs vulnerable to attacks, which is the motivation for our focus on this type of model.

When the system detects an attack on the external GPS, it adapts by excluding the compromised sensor data from the pose estimation process. In our setting, due to the internal placement and onboard feature, the IMU and camera remain reliable sensors that are less likely to be compromised. The resilient pose estimation, depending on the sensor attacked, is given by the unified expression:
\begin{equation}
    \mathcal{X} = \text{Fuse}(\text{camera}, \text{IMU}).
\end{equation}

By dynamically adjusting the sensor fusion strategy based on real-time threat analysis, the UAV can maintain good operational standards and accurate navigational capabilities even in hostile settings. This proactive and intelligent sensor fusion approach ensures that essential UAV operations can continue with minimum interruption, safeguarding both the mission and the hardware from potential threats.

\section{Experiments and Results}

\subsection{System Modeling}
\label{subsec:system_modeling}
\textbf{UAV Model}\hspace{5pt} 
For different states, inputs and measurements, we build two
representative system models according to Section~\ref{sec:residue_generator}. The first quadrotor model (Model I) has a good gyroscope and accelerometer. We use EKF to fuse information from GPS and IMU for state estimation. The system state is defined as:

\begin{equation}
x=\left[ \begin{array}{l}
	x_1\\
	x_2\\
	x_3\\
	x_4\\
	x_5\\
\end{array} \right] =\left[ \begin{array}{c}
	\mathbf{p}\\
	\mathbf{q}\\
	\dot{\mathbf{p}}\\
	\mathbf{b}_g\\
	\mathbf{b}_{\boldsymbol{a}}\\
\end{array} \right] \in \mathbb{R}^{15\times 1},
\label{eq:model1state}
\end{equation}
where $\mathbf{p} \in \mathbb{R}^{3\times 1}$ is position, $\mathbf{q} \in \mathbb{R}^{3\times 1}$ is orientation in \textit{SO}(3), $\dot{\mathbf{p}} \in \mathbb{R}^{3\times 1}$ is linear velocity, $\mathbf{b}_g \in \mathbb{R}^{3\times 1}$ is gyroscope bias and $\mathbf{b}_{\boldsymbol{a}} \in \mathbb{R}^{3\times 1}$ is accelerometer bias. The state-transition model can be written as:

\begin{equation}
    \dot{x}=\left[ \begin{array}{c}
	x_3\\
	\boldsymbol{G}\left( x_2 \right) ^{-1}\left( \boldsymbol{\omega }_m-x_4-\mathbf{n}_g \right)\\
	\mathbf{g}+\boldsymbol{R}\left( x_2 \right) \left( \mathbf{a}_m-x_5-\mathbf{n}_a \right)\\
	\mathbf{n}_{bg}\\
	\mathbf{n}_{ba}\\
\end{array} \right] + w,
\label{eq:model1f}
\end{equation}
where $\mathbf{g}$ is the gravity acceleration, $w \in \mathbb{R} ^{15\times 1}$ is process noise vector, $\boldsymbol{G}(x_2)\in \mathbb{R} ^{3\times 3}$ is the transformation from an Euler angular velocity to a body angular velocity vector, $\boldsymbol{R}\in \mathbb{R} ^{3\times 3}$ is the rotation matrix, $\mathbf{n}_g \in \mathbb{R} ^{3\times 1}$, $\mathbf{n}_a \in \mathbb{R} ^{3\times 1}$, $\mathbf{n}_{bg} \in \mathbb{R} ^{3\times 1} =\dot{\mathbf{b}}_g$ and $\mathbf{n}_{ba} \in \mathbb{R} ^{3\times 1} =\dot{\mathbf{b}}_a$ are the noise terms inside IMU, $\boldsymbol{\omega }_m\in \mathbb{R} ^{3\times 1}$ and $\mathbf{a}_m \in \mathbb{R} ^{3\times 1}$ are the estimate of angular velocity and linear acceleration, respectively. The observation model can be written as:

\begin{equation}
    z=\left[ \begin{matrix}
    	I&		0&		0&		0&		0\\
    	0&		I&		0&		0&		0\\
    \end{matrix} \right] \left[ \begin{array}{c}
    	\mathbf{p}\\
    	\mathbf{q}\\
    	\dot{\mathbf{p}}\\
    	\mathbf{b}_g\\
    	\mathbf{b}_{\boldsymbol{a}}\\
    \end{array} \right] + v.
\label{eq:model1g}
\end{equation}
where $v \in \mathbb{R} ^{15\times 1}$ is measurement noise vector.

The second quadrotor model (Model II) uses more sensor information (stereo camera) and leverages augmented states from previous keyframes. EKF will integrate information from GPS, IMU, and visual odometry (obtained from a stereo camera with optical flow calculation) for state estimation. The system state is defined as: 

\begin{equation}
\dot{x}=\left[ \begin{array}{c}
	x_1\\
	x_2\\
	x_3\\
	x_4\\
	x_5\\
	x_{\mathrm{F}_1}\\
	x_{\mathrm{F}_2}\\
\end{array} \right] =\left[ \begin{array}{c}
	\mathbf{p}\\
	\mathbf{q}\\
	\dot{\mathbf{p}}\\
	\mathbf{b}_g\\
	\mathbf{b}_a\\
	\mathbf{p}_{\mathrm{F}}\\
	\mathbf{q}_{\mathrm{F}}\\
\end{array} \right] \in \mathbb{R}^{21\times 1}.
\label{eq:model2state}
\end{equation}
The state-transition model can be written as:
\begin{equation}
    \dot{x}=\left[ \begin{array}{c}
	x_3\\
	\boldsymbol{G}\left( x_2 \right) ^{-1}\left( \boldsymbol{\omega }_m-x_4-\mathbf{n}_g \right)\\
	\mathbf{g}+\boldsymbol{R}\left( x_2 \right) \left( \mathbf{a}_m-x_5-\mathbf{n}_a \right)\\
	\mathbf{n}_{bg}\\
	\mathbf{n}_{ba}\\
	0\\
	0\\
\end{array} \right] + w.
\label{eq:model2f}
\end{equation}
The observation model can be written as:
\begin{equation}
z=\left[ \begin{array}{c}
	\boldsymbol{R}\left( \mathbf{q}_{\mathrm{F}} \right) ^T\left( \mathbf{p}-\mathbf{p}_{\mathrm{F}} \right)\\
	\mathrm{Euler}\left( \boldsymbol{R}\left( \mathbf{q}_{\mathrm{F}} \right) ^T\boldsymbol{R}(\mathbf{q}) \right)\\
\end{array} \right] + v.
\label{eq:model2g}
\end{equation}
These two UAV models are simplified versions of commonly used models in \cite{mourikis2007multi}. We show that the above two UAV models satisfy the conditions in Theorem 1. Due to the continuity of UAV's transition dynamics, $\left\| \nabla _xf(\hat{x}_{k|k}) \right\| \leqslant \beta _1, \left\| \nabla _xg(\hat{x}_{k|k}) \right\| \leqslant \beta _2$ hold as long as $\hat{x}_{k|k}$ is bounded. Meanwhile, $P_{k|k}$ is bounded otherwise the EKF will fail to work. Therefore condition 1) is satisfied in normal situations. As the UAV dynamics are discretized using Euler method in practice,  thus we have
\begin{align*}
    x_{k+1} = f(x_k,u_k) = x(t) + dt*\phi(x(t),u(t)) ~~\text{as}~~dt\rightarrow 0^+.
\end{align*}
where $\dot{x}(t) = \phi(x(t),u(t))$ denotes the continuous-time dynamics. Notice that $\nabla_x f(\cdot) = I + dt*\nabla_x\phi(x(t),u(t))$ is always non-singular for a sufficiently small $dt$. Therefore condition 2) is satisfied. The condition 3) is equivalent to 
{\small\begin{align*}
   &\| f(x_k) - f(\hat{x}_{k|k})-\nabla_x f(\hat{x}_{k|k})(x_k-\hat{x}_{k|k})\|\le \epsilon _{\varphi}\left\| x_k-\hat{x}_{k|k} \right\| ^2,\\
   &\|g(x_k) - g(\hat{x}_{k|k}) - \nabla_x g(\hat{x}_{k|k}) (x_k-\hat{x}_{k|k})\| \le \epsilon _{\chi}\left\| x_k-\hat{x}_{k|k} \right\| ^2.
\end{align*}}

\noindent
The inequalities are guaranteed by the Lipschitz continuity of $\nabla_x f(\cdot)$ and $\nabla_x g(\cdot)$, which can be verified numerically.

\vspace{5pt}

\textbf{Noise Model}\hspace{5pt} 
In this paper, we consider two representative types of non-Gaussian white noise distributions in the state-transition and observation model: Exponential and Laplace~\cite{carrillo2014state}. Each type of noise exhibits distinct statistical properties that can significantly affect the system's performance. 

For exponential noise, each component $n_{e}(i)$ is characterized by an exponential distribution with the PDF:
\begin{equation}
f_{n_e}(i|\kappa )=\kappa e^{-\kappa i},\quad i\ge 0,
\end{equation}
where $\kappa$ is the rate parameter.
For Laplacian noise, the distribution for noise component $n_{l}(i)$ is given by the PDF:
\begin{equation}
f_{n_l}(i|\mu ,b)=\frac{1}{2b}e^{\left( -\frac{|i-\mu |}{b} \right)},
\end{equation}
where $\mu$ is the location parameter, typically set to 0, and $b > 0$ determines the spread of the distribution.

\begin{figure}[h]
    \centering
    \hfill 
    \begin{subfigure}[b]{0.49\columnwidth}
        \includegraphics[width=\linewidth]{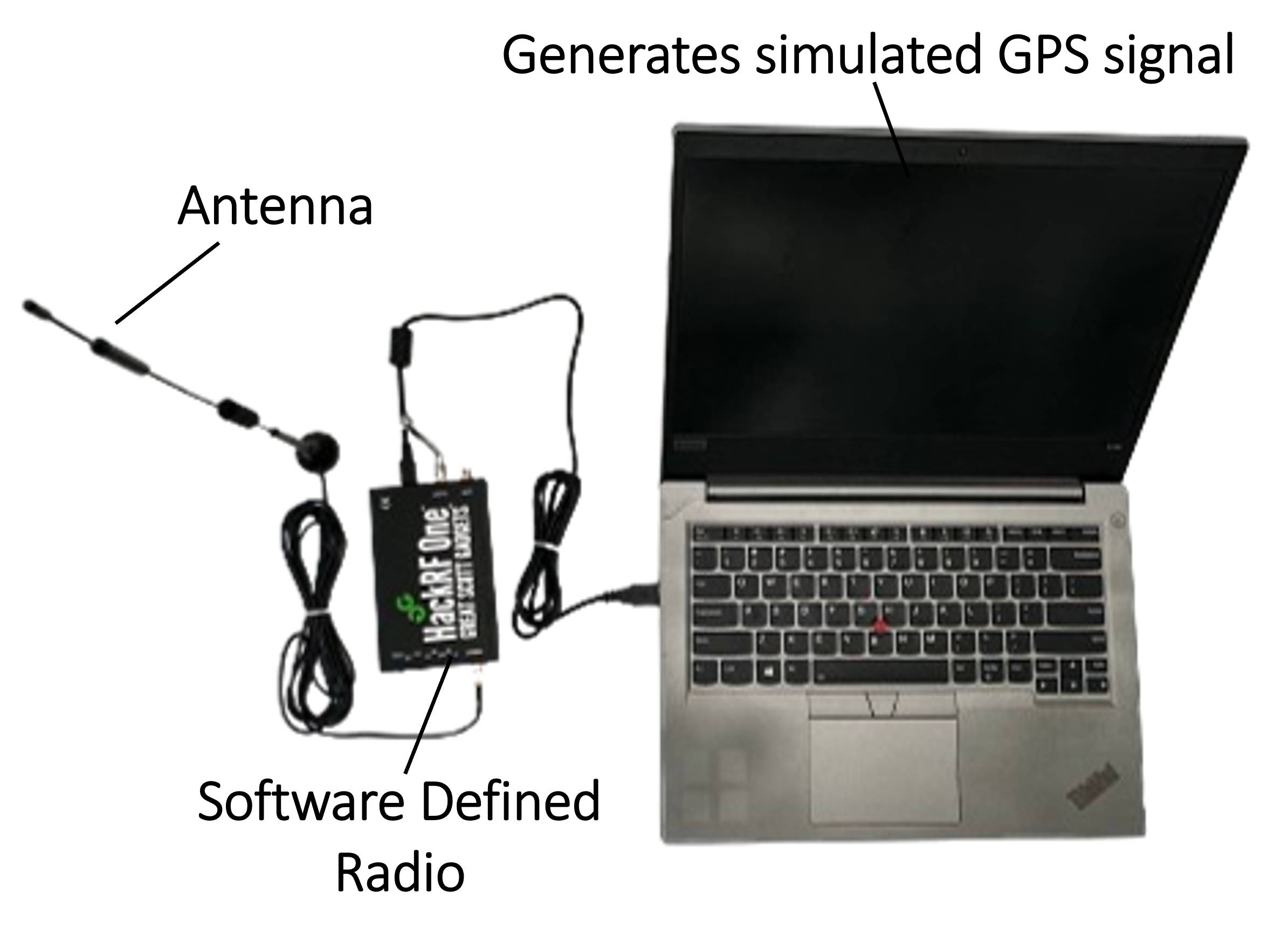}
        \caption{Real attacks launcher.}
        \label{fig:attack_signal}
    \end{subfigure}
    \hfill
    \begin{subfigure}[b]{0.49\columnwidth}
        \includegraphics[width=\linewidth]{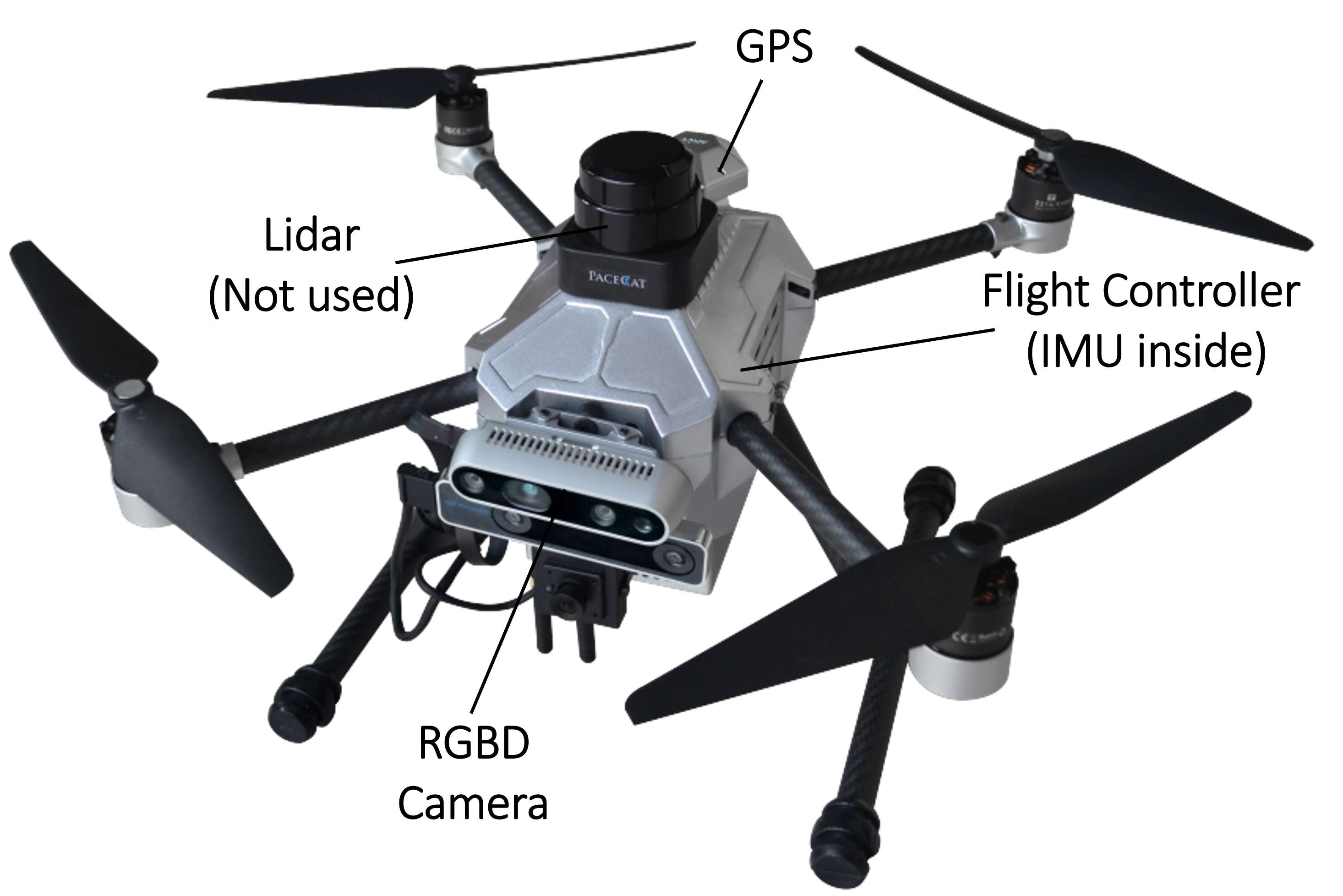}
        \caption{Quadrotor UAV system.}
        \label{fig:uav}
    \end{subfigure}
    \caption{Real-world experiment setup.}
    \label{fig:hardware_combined}
\end{figure}

\subsection{Attack Scenario and Training Details}

The external GPS may be subject to false data injection attacks, causing bias in sensor readings at certain discrete moments or consecutive periods, thereby deceiving the whole control system. Compared to the DoS attacks and reply attacks, FDI attacks are more common and more deceptive to detect. In our setting, sparse and continuous FDI attacks will be implemented on different models with different noises, and the magnitude of the attack will also have different sizes.

To train our model, we let the UAV fly over a wide and long range, and during the flight, we apply intermittent attacks of different durations and amplitudes on the GPS. It is worth noting that in addition to implementing simulation attacks through internal programs, we specifically implement electronic attacks against the physical UAV system, as shown in Fig.~\ref{fig:hardware_combined}. We collect training data containing 83,700 residue samples and test data containing 18,700 residue samples for subsequent semi-supervised training, of which approximately 12\% of the time steps are under attack.

\subsection{Simulation Experiment}

\begin{figure*}[!t]
  \centering
  \begin{subfigure}{0.32\textwidth}
    \centering
    \includegraphics[width=\textwidth]{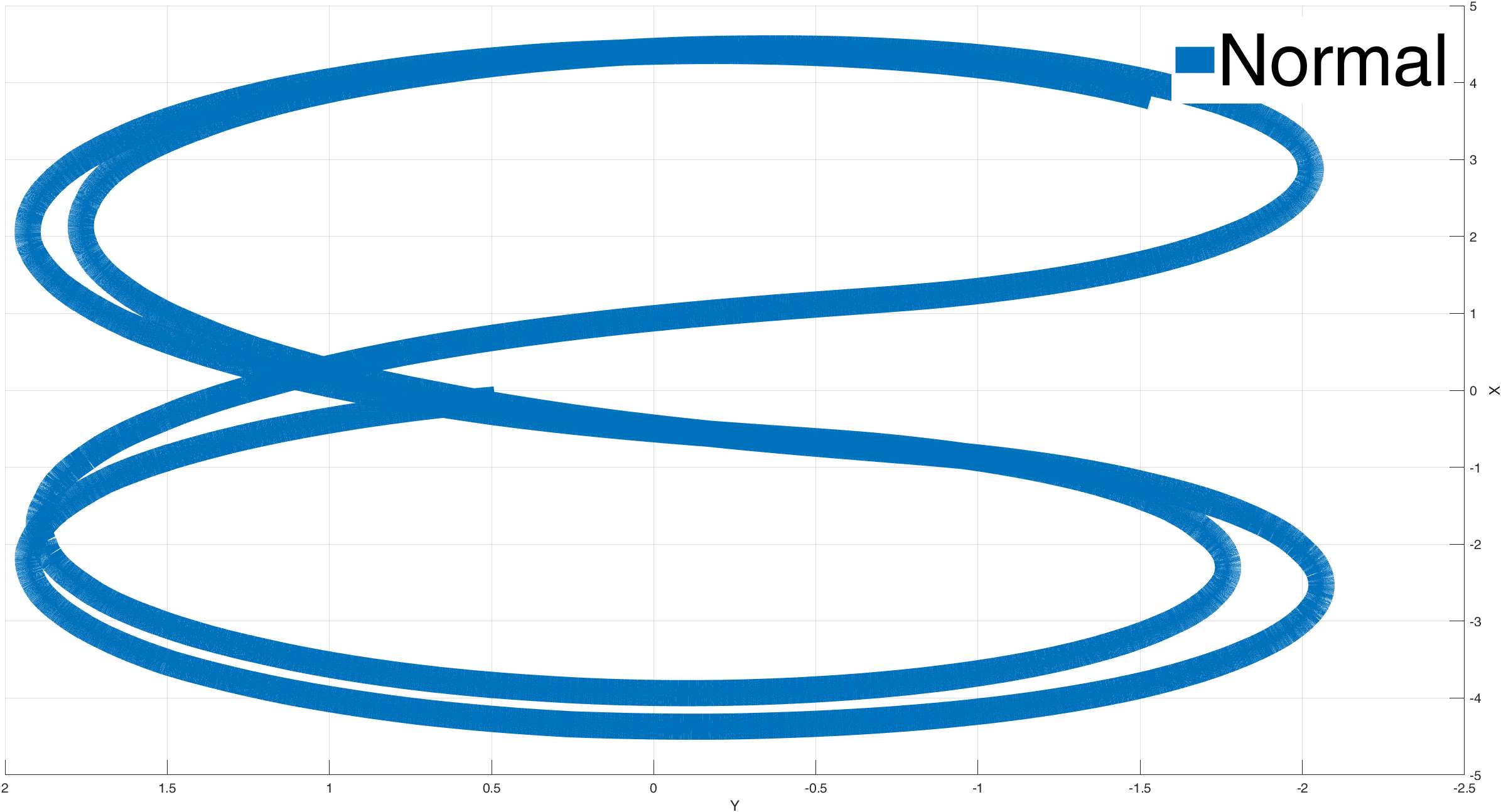}
    \caption{UAV with normal operation.}
    \label{fig:subfiga}
  \end{subfigure}
  \hfill
  \begin{subfigure}{0.32\textwidth}
    \centering
    \includegraphics[width=\textwidth]{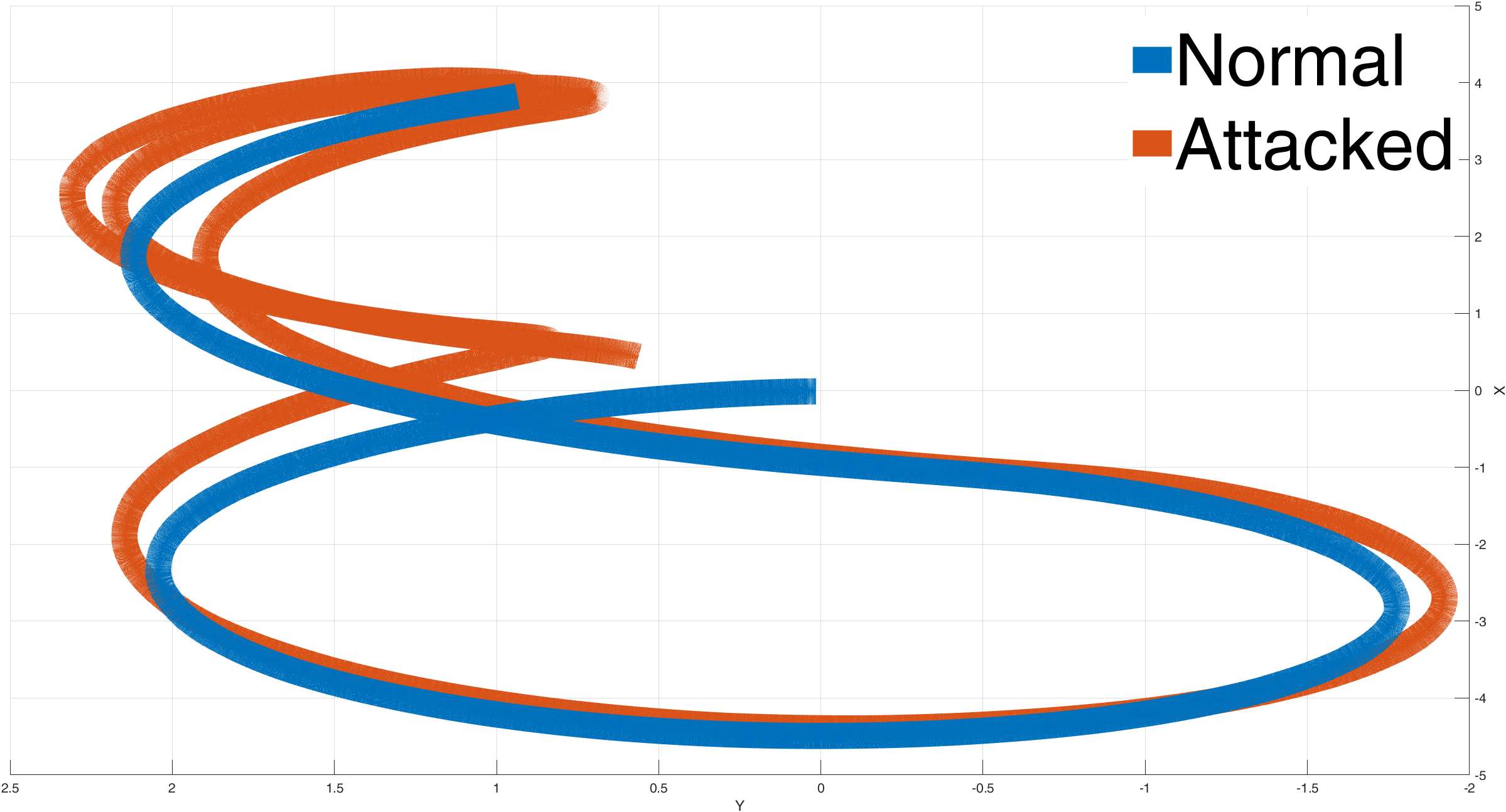}
    \caption{UAV under attack.}
    \label{fig:subfigb}
  \end{subfigure}
  \hfill
  \begin{subfigure}{0.32\textwidth}
    \centering
    \includegraphics[width=\textwidth]{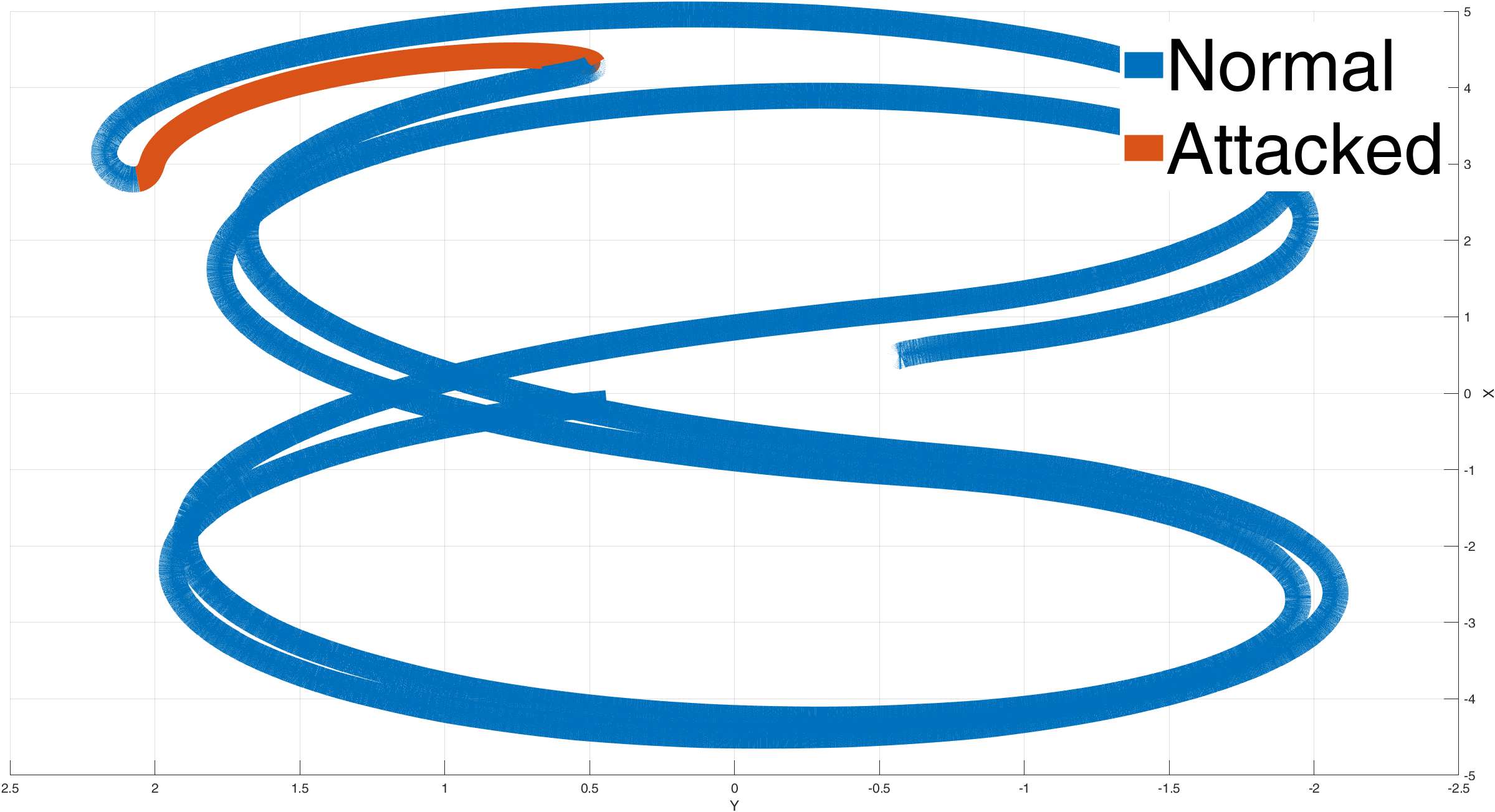}
    \caption{UAV under attack with detection.}
    \label{fig:subfigc}
  \end{subfigure}
  
  \caption{UAV autonomous ``Eight-shape'' flight in three different cases.}
  \label{fig:simu1}
\end{figure*}

\begin{figure*}[!t]
  \centering
  \begin{subfigure}{0.32\textwidth}
    \centering
    \includegraphics[width=\textwidth]{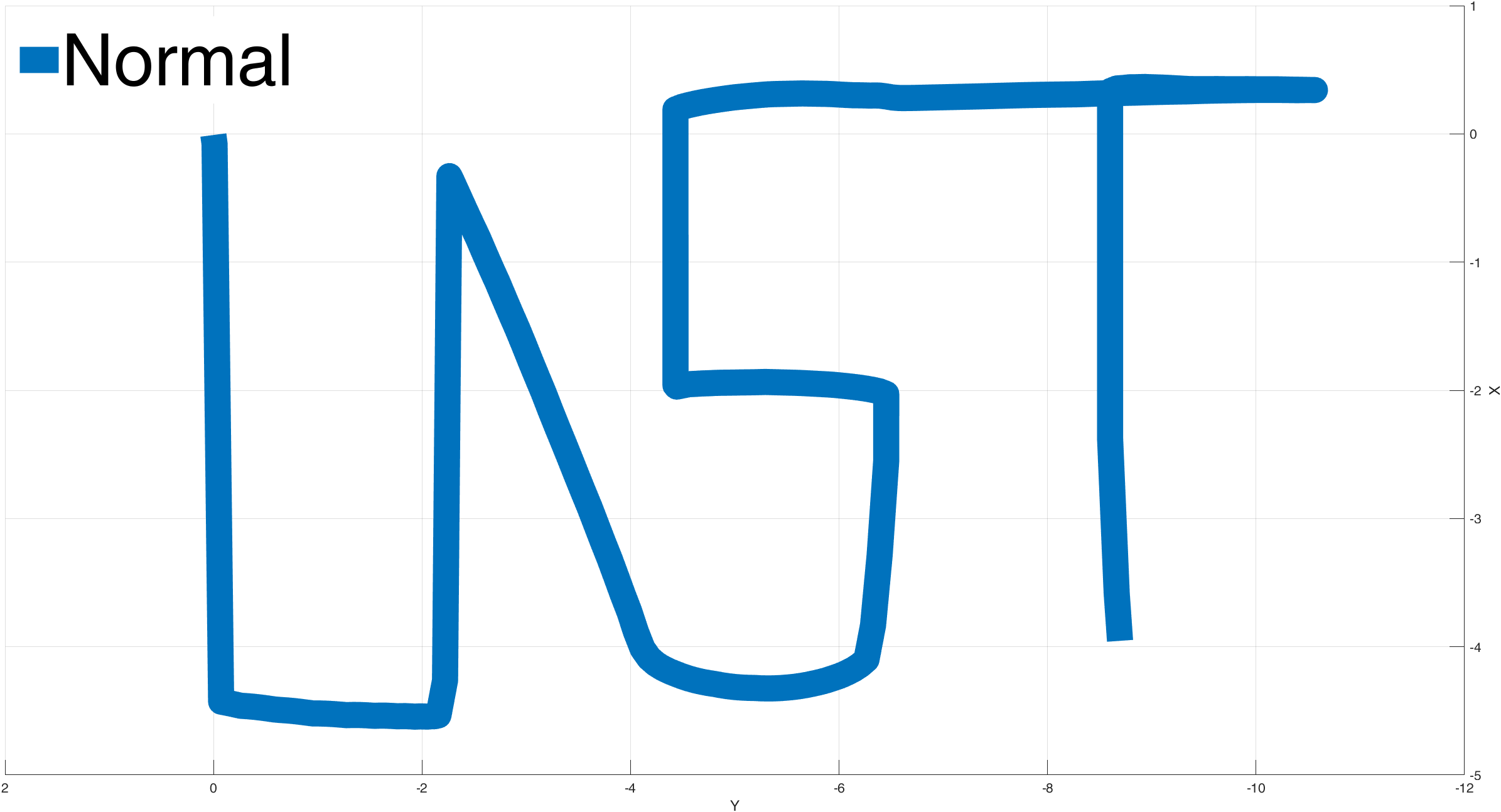}
    \caption{UAV with normal operation.}
    \label{fig:subfiga2}
  \end{subfigure}
  \hfill
  \begin{subfigure}{0.32\textwidth}
    \centering
    \includegraphics[width=\textwidth]{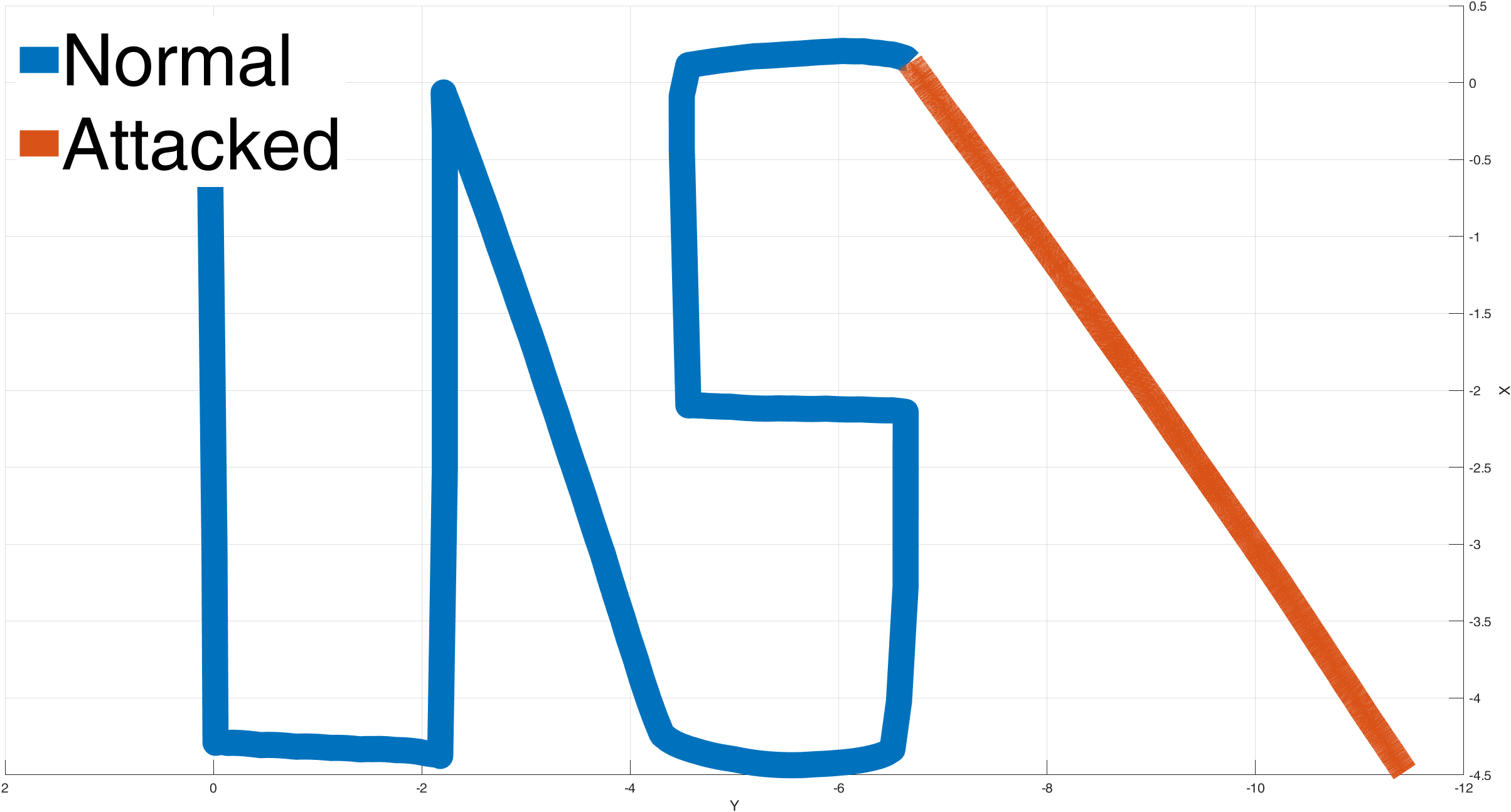}
    \caption{UAV under attack.}
    \label{fig:subfigb2}
  \end{subfigure}
  \hfill
  \begin{subfigure}{0.32\textwidth}
    \centering
    \includegraphics[width=\textwidth]{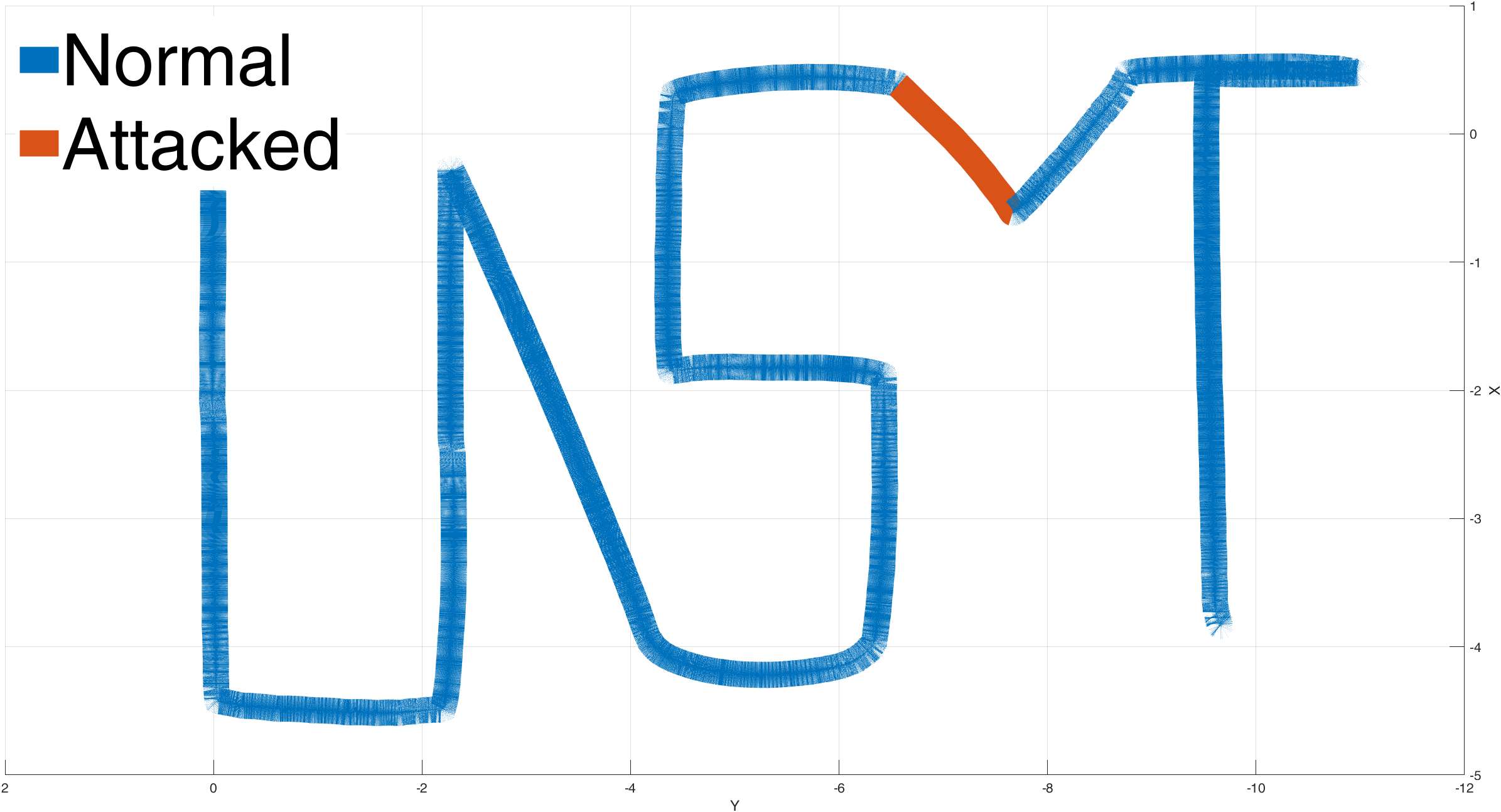}
    \caption{UAV under attack with detection.}
    \label{fig:subfigc2}
  \end{subfigure}
  
  \caption{UAV autonomous ``UST-shape'' flight in three different cases.}
  \label{fig:simu2}
\end{figure*}

\begin{figure*}[!t]
  \centering
  \begin{subfigure}{0.32\textwidth}
    \centering
    \includegraphics[width=\textwidth]{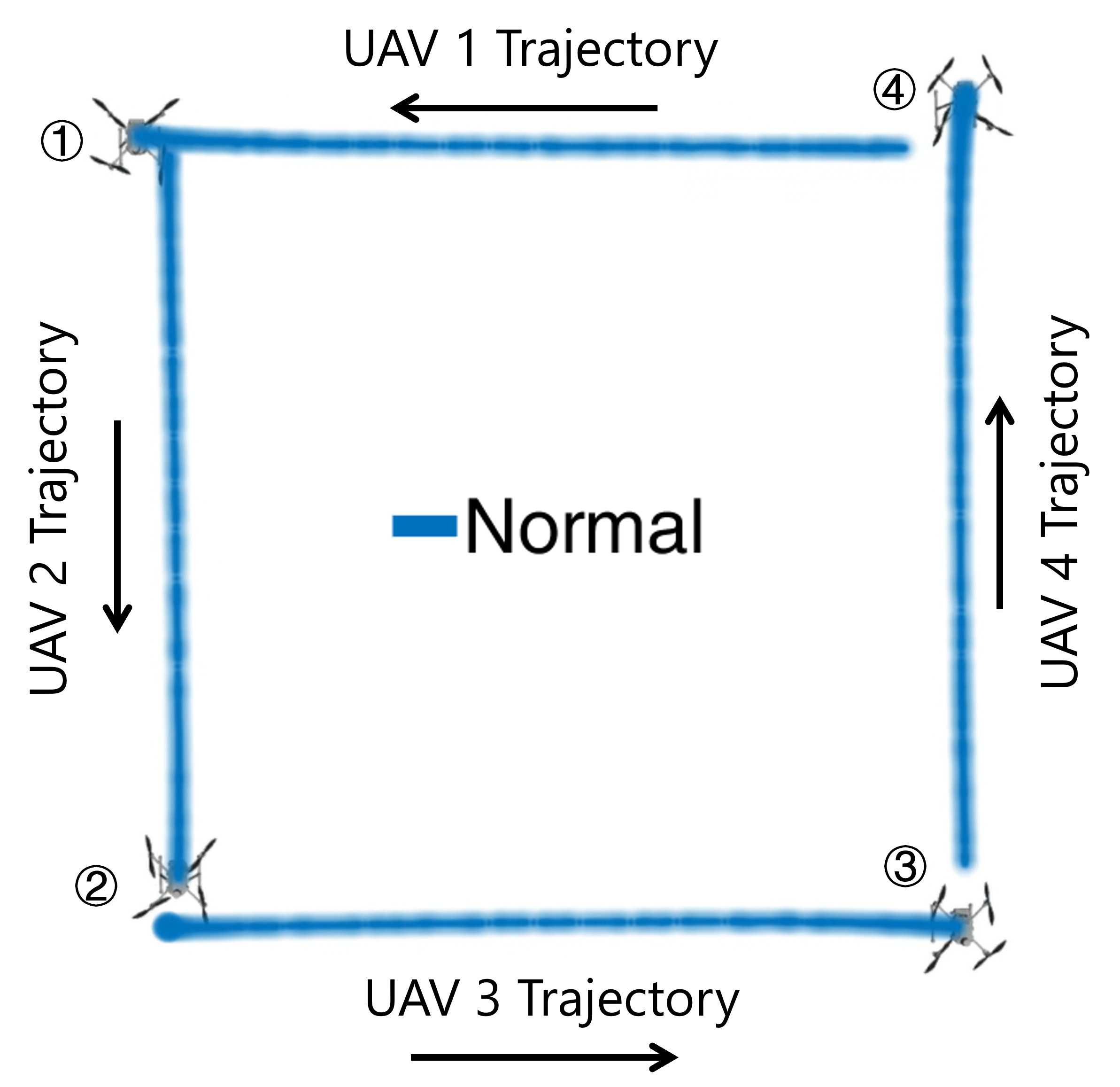}
    \caption{UAV with normal operation.}
    \label{fig:subfiga3}
  \end{subfigure}
  \hfill
  \begin{subfigure}{0.32\textwidth}
    \centering
    \includegraphics[width=\textwidth]{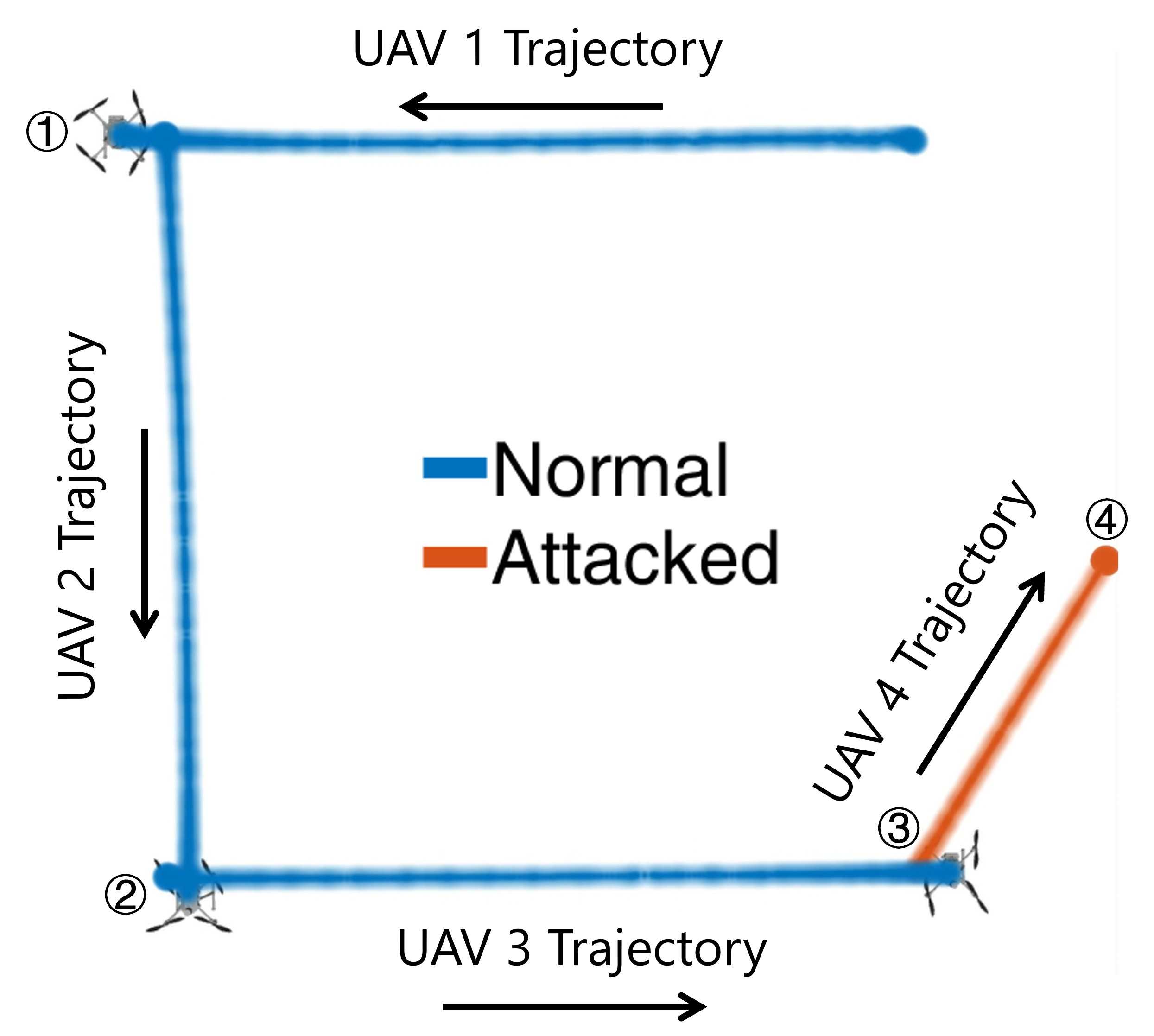}
    \caption{UAV under attack.}
    \label{fig:subfigb3}
  \end{subfigure}
  \hfill
  \begin{subfigure}{0.32\textwidth}
    \centering
    \includegraphics[width=\textwidth]{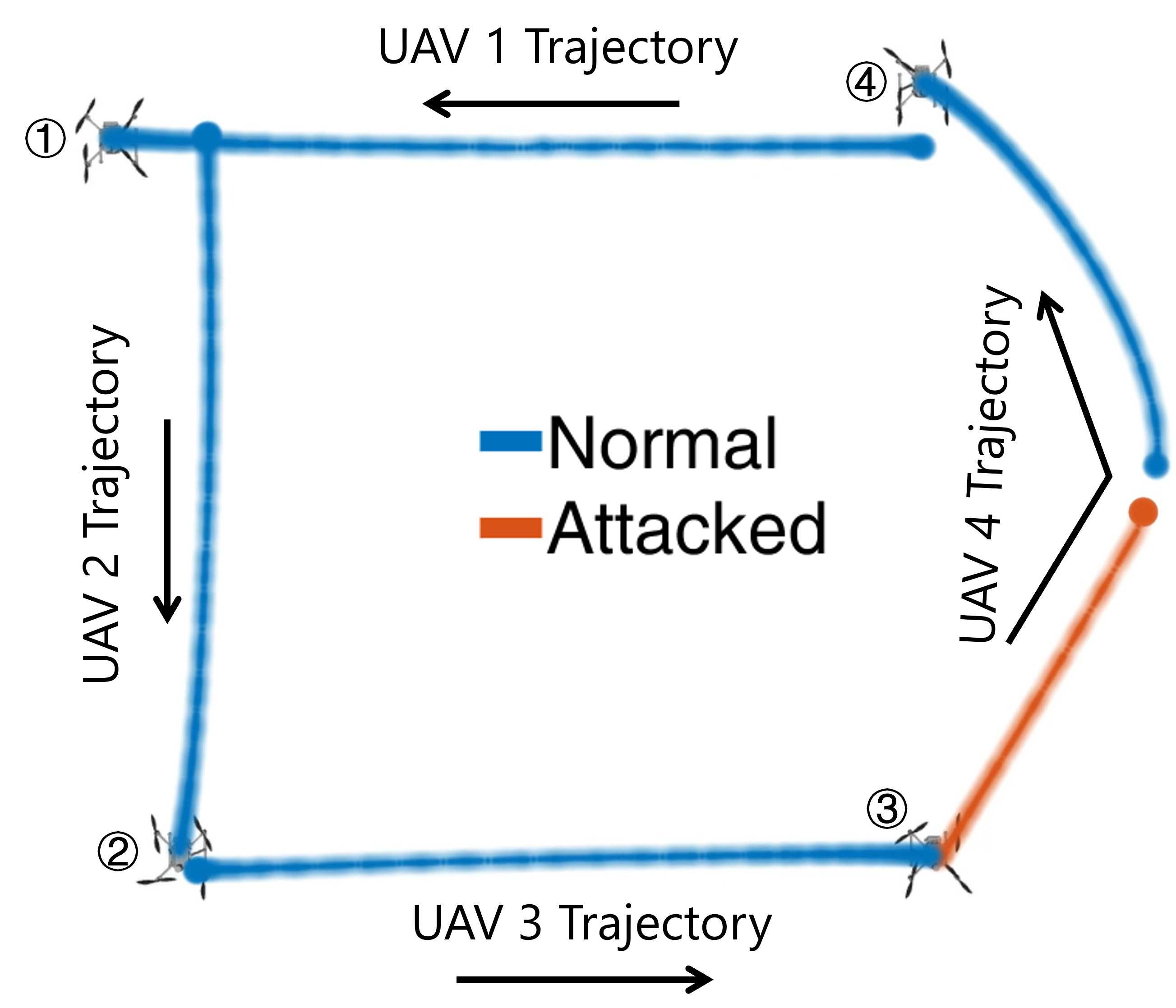}
    \caption{UAV under attack with detection.}
    \label{fig:subfigc3}
  \end{subfigure}
  
  \caption{Swarm UAV autonomous formation flight in three different cases.}
  \label{fig:simu3}
\end{figure*}

We build a simulation platform based on ROS/Gazebo, which enables UAVs to autonomously fly along specific pre-planned trajectories. Autonomous flight in each scenario includes three cases: the UAV flying normally, the UAV subjected to a GPS spoofing attack without an attack detection framework deployed, and the UAV under a GPS spoofing attack but with our attack detection framework in place. In the first single-UAV simulation scenario, the UAV takes off from a central location and is expected to complete two ``Eight-shape'' circles, with the attack occurring in the second half of the first circle, as shown in Fig.~\ref{fig:simu1}. In the second single-UAV simulation scenario, the UAV takes off from the upper left corner and completes a ``UST-shape" flight to the right, with the attack occurring after completing the letter ``S" trajectory, as shown in Fig.~\ref{fig:simu2}. Moreover, we further conduct a formation flight consisting of four UAVs, one of which is subjected to a GPS spoofing attack, leading to the disruption of the cluster formation. Specifically, The four UAVs are first located at the four vertices of the square, and then the four UAVs fly counterclockwise to the next vertex respectively, and the UAV numbered 3 will be attacked, as shown in Fig.~\ref{fig:simu3}. It is important to note that the time, duration and range of the external attack are totally unknown to UAVs. 

\begin{figure*}[!t]
  \centering
  \begin{subfigure}{0.32\textwidth}
    \centering
    \includegraphics[width=\textwidth]{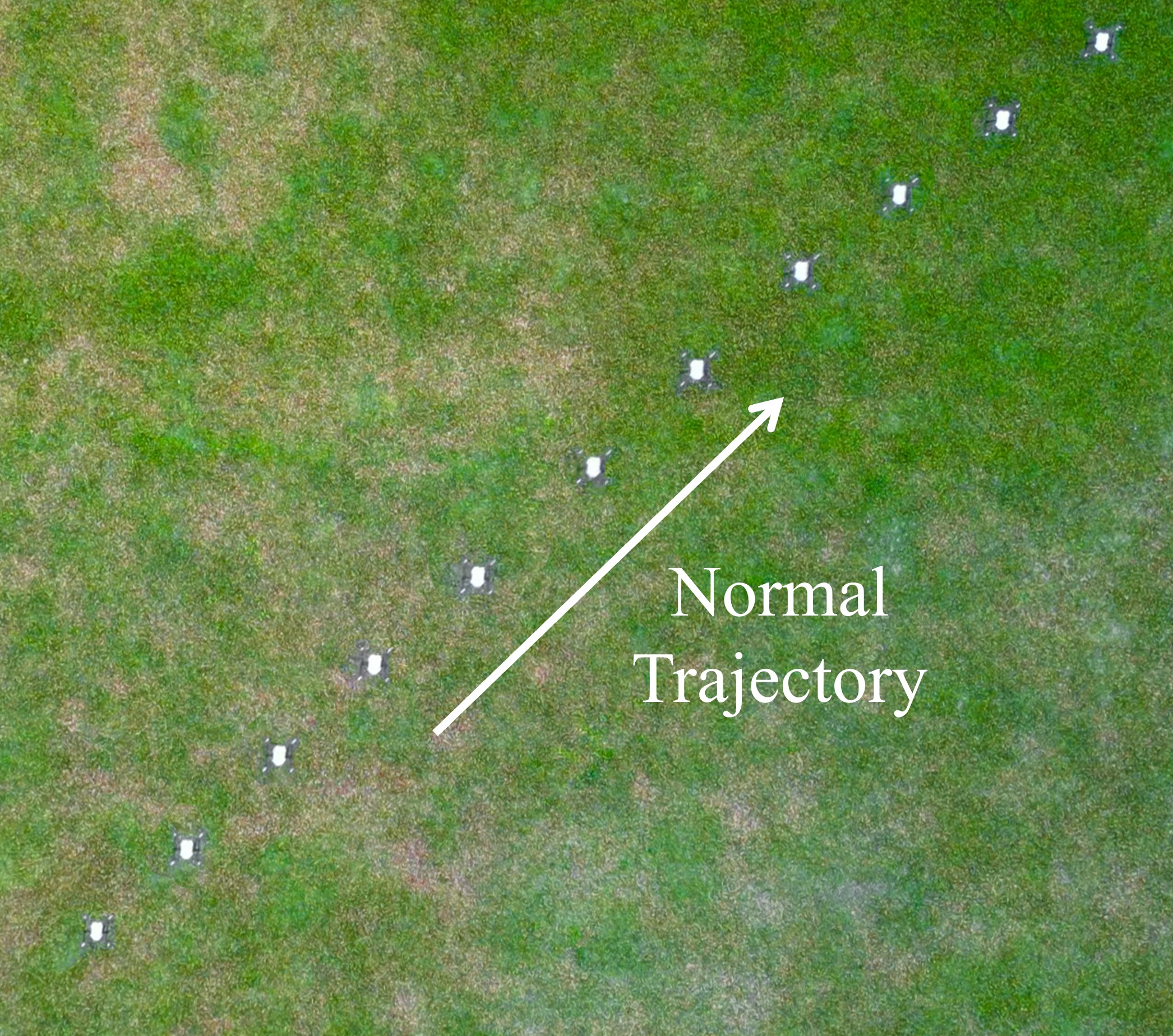}
    \caption{UAV with normal operation.}
    \label{fig:subfiga4}
  \end{subfigure}
  \hfill
  \begin{subfigure}{0.32\textwidth}
    \centering
    \includegraphics[width=\textwidth]{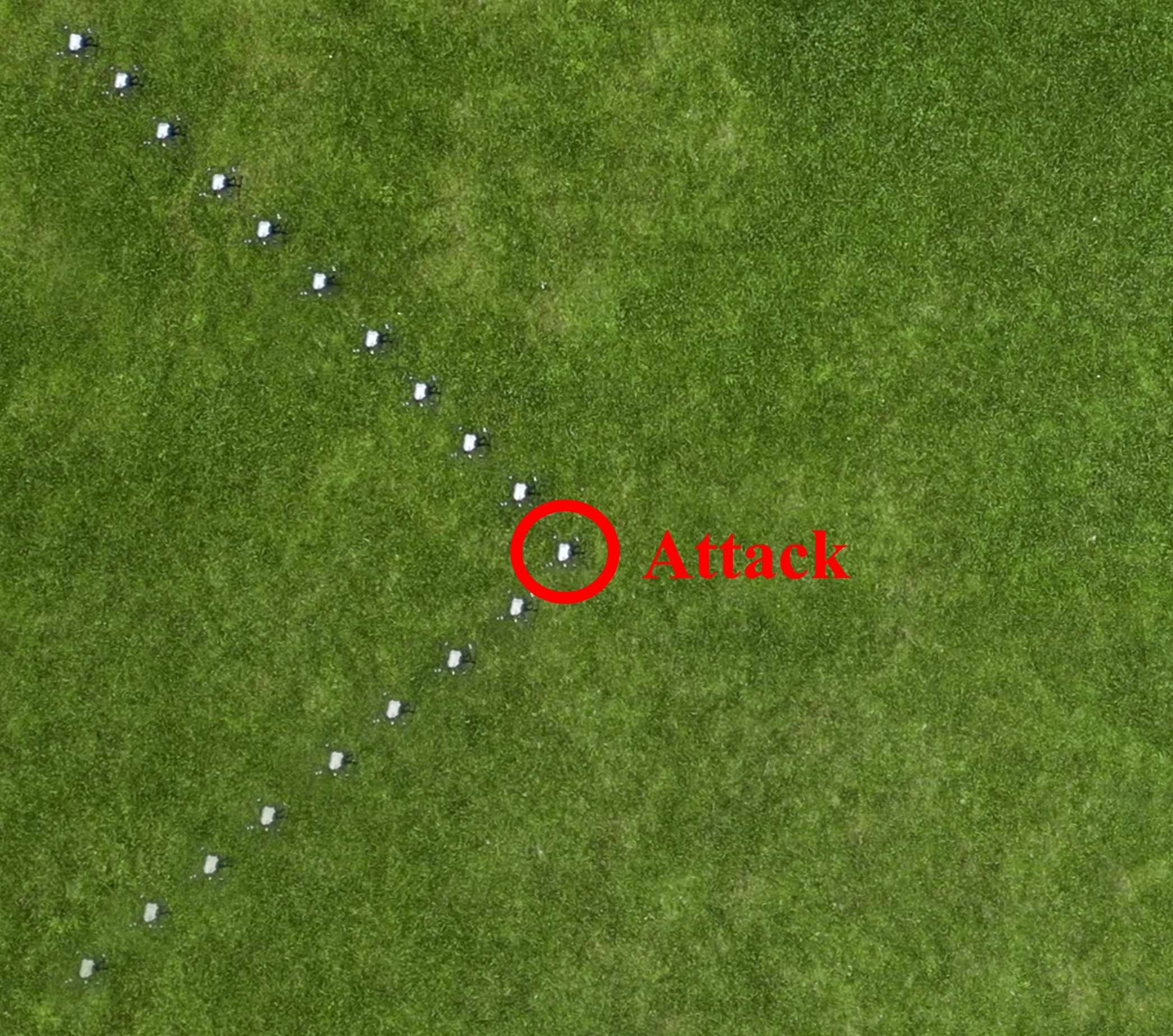}
    \caption{UAV under attack.}
    \label{fig:subfigb4}
  \end{subfigure}
  \hfill
  \begin{subfigure}{0.32\textwidth}
    \centering
    \includegraphics[width=\textwidth]{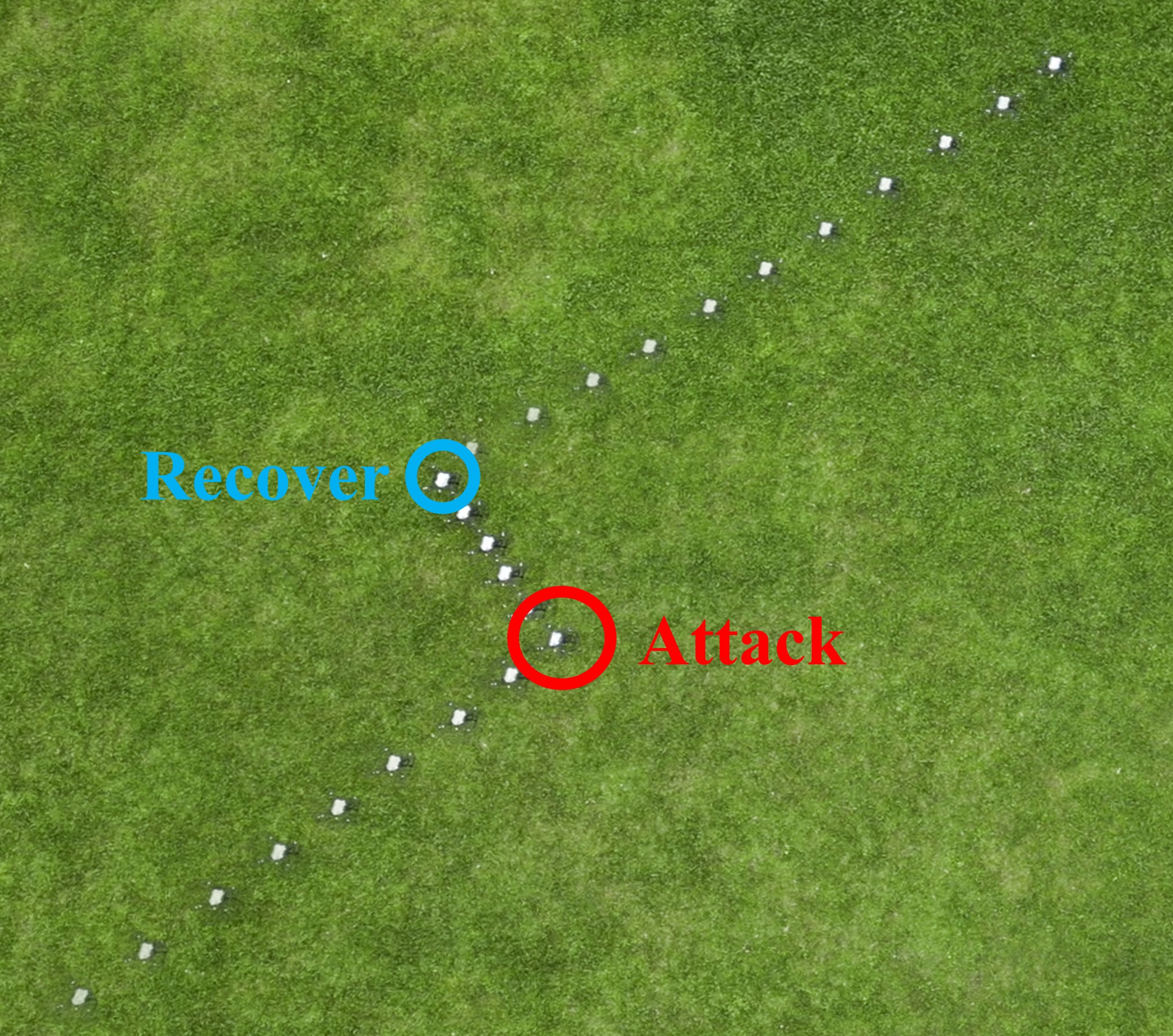}
    \caption{UAV under attack with detection.}
    \label{fig:subfigc4}
  \end{subfigure}
  
  \caption{UAV real autonomous ``diagonal'' flight in three different cases.}
  \label{fig:real1}
\end{figure*}

\begin{figure*}[!t]
  \centering
  \begin{subfigure}{0.32\textwidth}
    \centering
    \includegraphics[width=\textwidth]{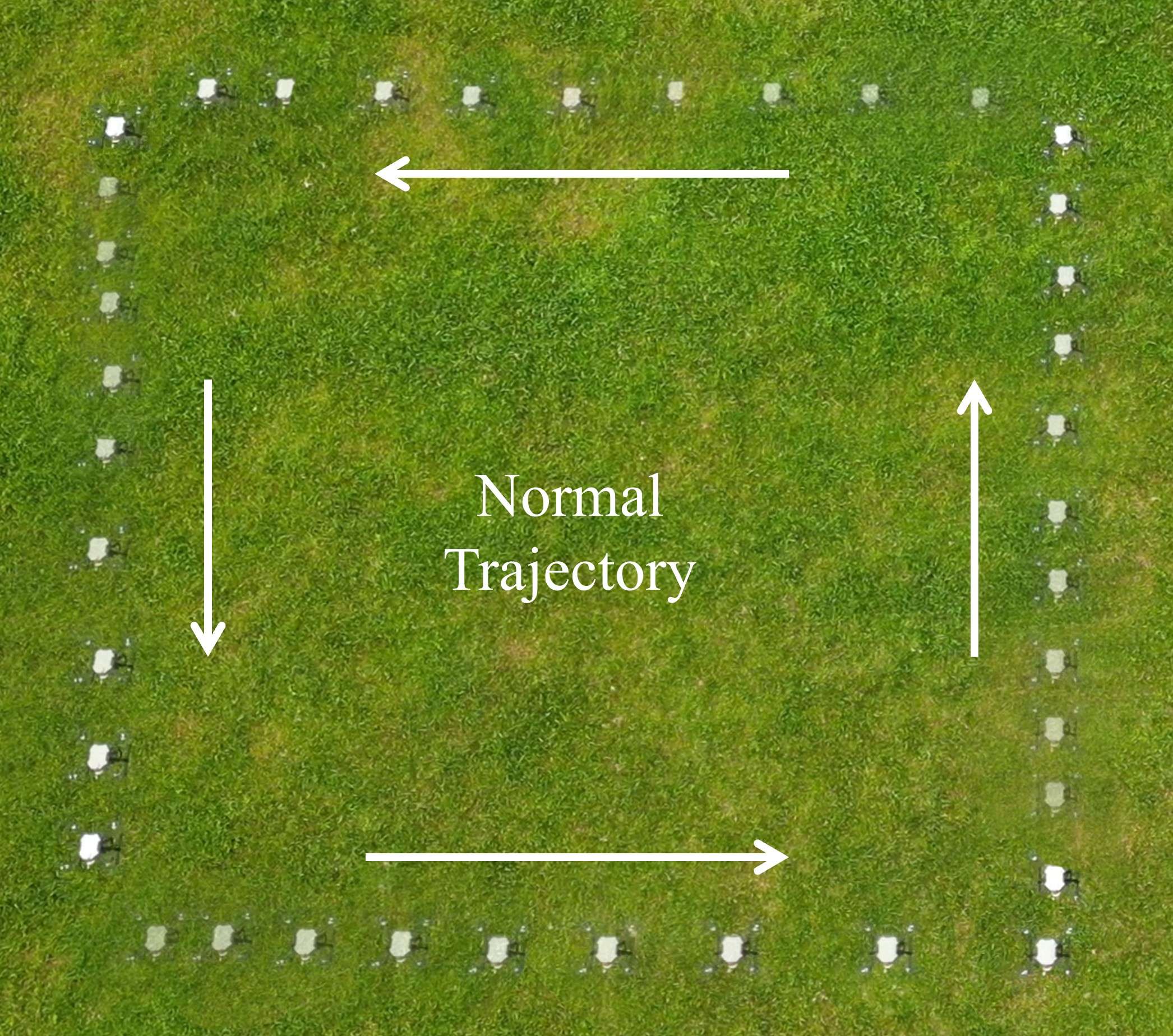}
    \caption{UAV with normal operation.}
    \label{fig:subfiga5}
  \end{subfigure}
  \hfill
  \begin{subfigure}{0.32\textwidth}
    \centering
    \includegraphics[width=\textwidth]{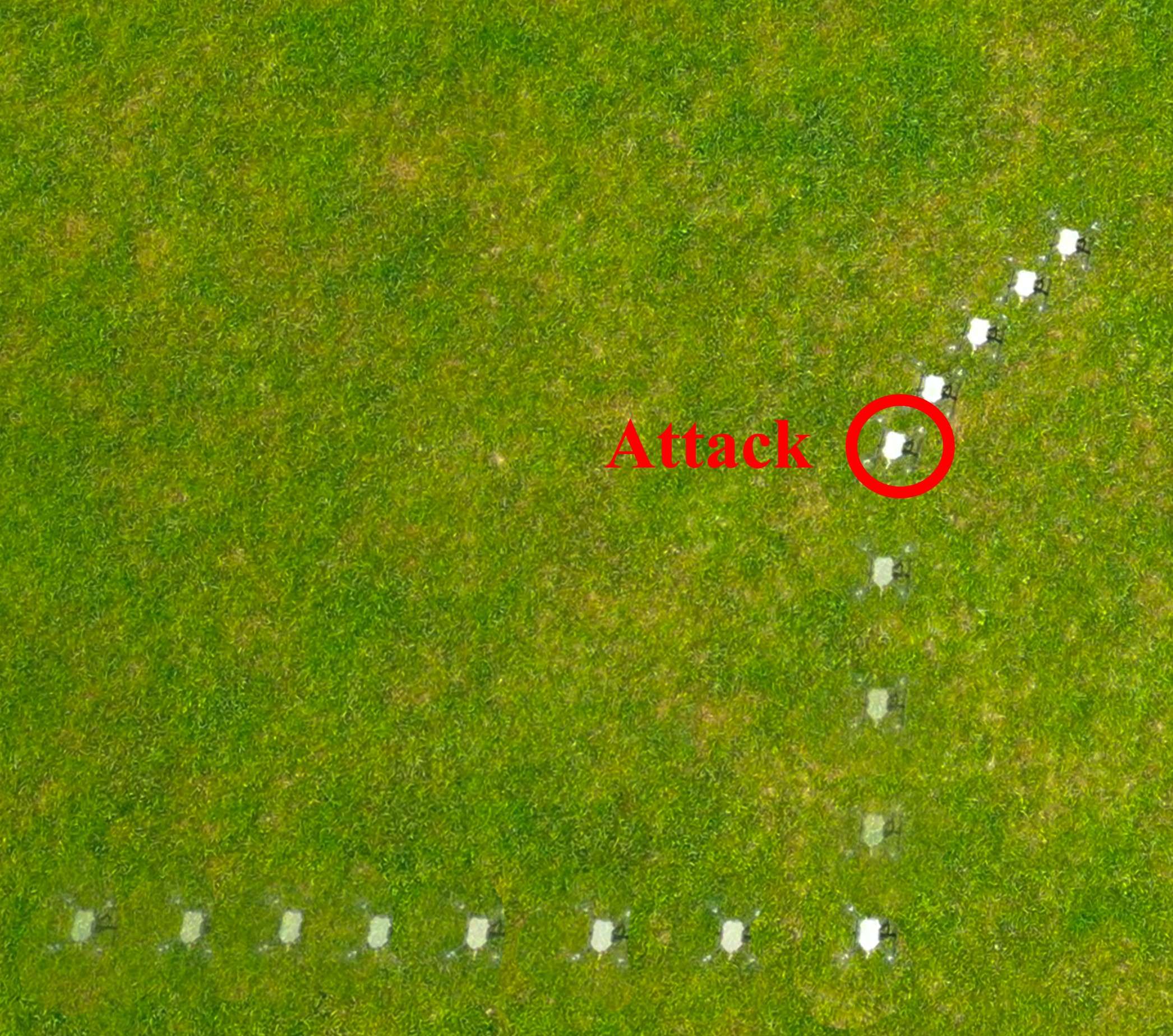}
    \caption{UAV under attack.}
    \label{fig:subfigb5}
  \end{subfigure}
  \hfill
  \begin{subfigure}{0.32\textwidth}
    \centering
    \includegraphics[width=\textwidth]{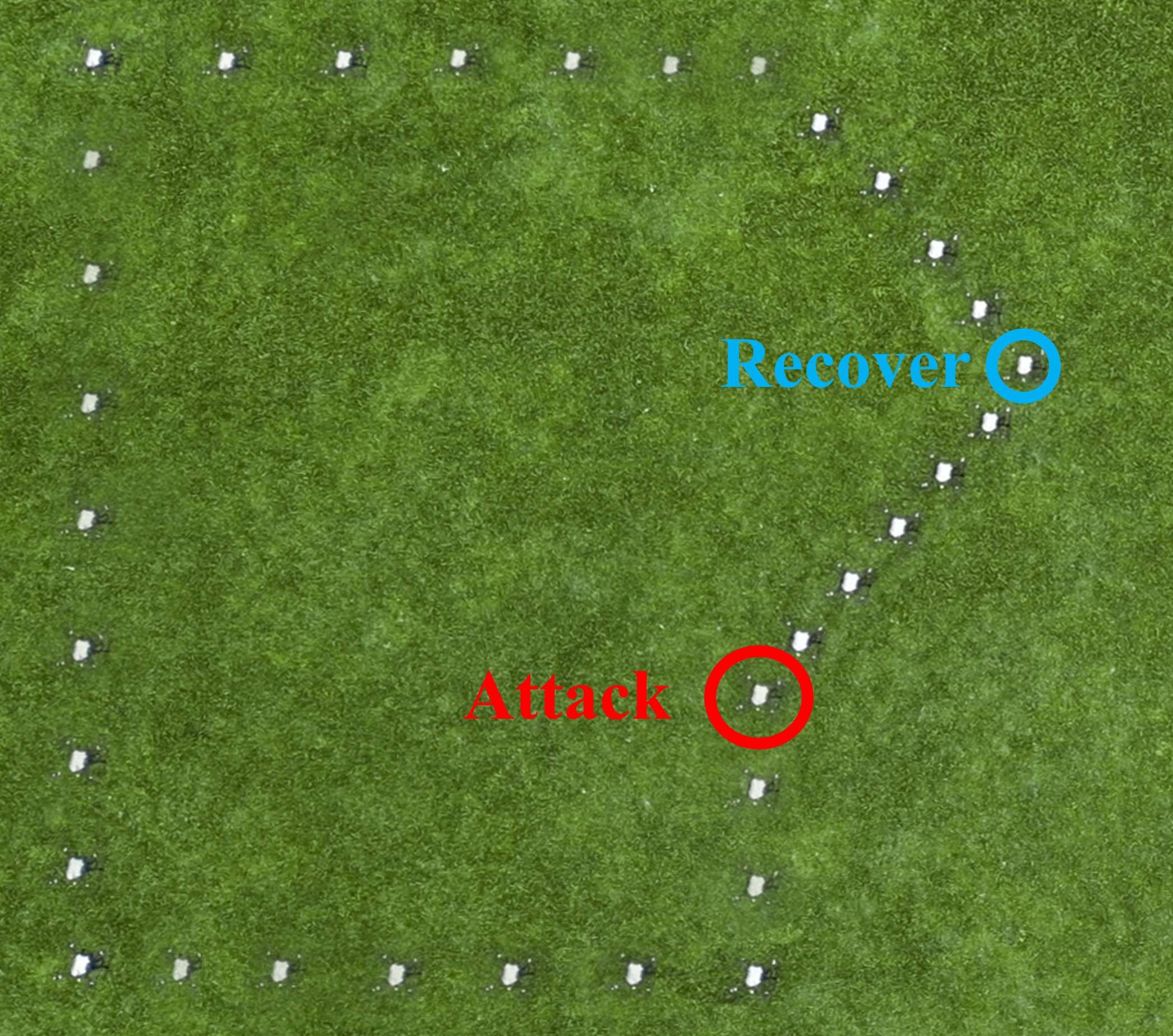}
    \caption{UAV under attack with detection.}
    \label{fig:subfigc5}
  \end{subfigure}
  
  \caption{UAV real autonomous ``square'' flight in three different cases.}
  \label{fig:real2}
\end{figure*}

\subsection{Real-world Experiment}

Furthermore, We deploy the related methods on a real UAV and conduct extensive experiments outdoors, which is rare in existing research. To carry out a realistic FDI attack, we utilize a laptop to generate fake GPS signals and broadcast them using a HackRF One Radio\footnote{https://hackrfone.com/} equipped with an antenna, as shown in Fig.~\ref{fig:attack_signal}. Our UAV system is a PX4-based\footnote{https://px4.io/} quadrotor UAV equipped with different sensors, including an external localization GPS receiver, a D435i RGBD stereo camera\footnote{https://www.intelrealsense.com/depth-camera-d435i} and an IMU, as shown in Fig.~\ref{fig:uav} and other sensors onboard are not used in our experiments. During the experiment on campus, we strictly abide by local laws and regulations and the range of our GPS jammer only affects our own aircraft. Similar to simulation, autonomous flight in each scenario also includes three cases. In the first real-world scenario, the UAV takes off from the lower left corner and flies diagonally to the upper right corner, and attacks will occur midway through the flight, as shown in Fig.~\ref{fig:real1}. In the second real-world scenario, the UAV takes off from the lower left corner, draws a square trajectory counterclockwise, and finally returns to the starting point. The attack will occur when the UAV flies to the second side of the square, as shown in Fig.~\ref{fig:real2}. As with the simulation, the specifics of the UAV’s attack are completely unknown in advance. Whether in simulation or in the real world, when we deploy the model to the edge, the inference time of the model is about 1 second, which means that the UAV will use incorrect position information for pose estimation, but due to the short duration, it is totally acceptable in our scenario. Naturally, choosing more advanced edge computing devices and accelerating hardware computing can further reduce the response time.

\subsection{Comparison Result and Analysis}

\begin{figure}[h]
    \centering
    \includegraphics[width=\linewidth]{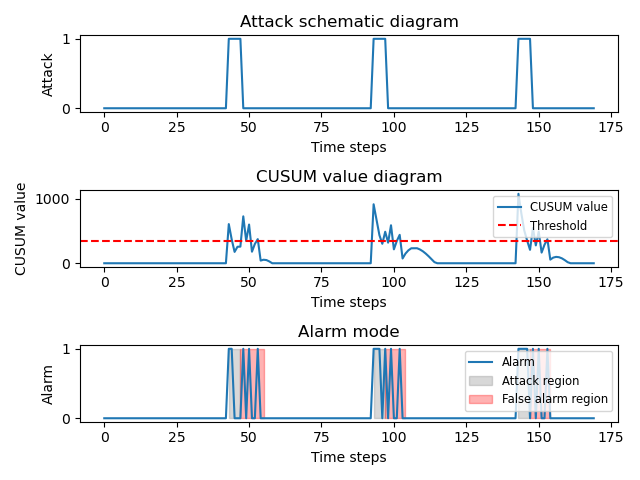}
    \caption{CUSUM experiment results.}
    \label{fig:CUSUM_3_plots}
\end{figure}

We reproduce the most commonly used attack detection methods based on traditional statistics~\cite{yoon2019towards}~\cite{liu2019secure}~\cite{muniraj2017framework} and machine learning~\cite{shafique2021detecting}~\cite{khoei2022comparative}~\cite{wu2023highly}. It can be seen from the experiments that our method has outstanding performance under different models, noises and attack amplitudes, and it is better than many traditional methods and learning-based methods in most indicators. 

For the analysis of traditional state-of-the-art methods, we have selected CUSUM~\cite{yoon2019towards}\cite{liu2019secure}, SPRT~\cite{muniraj2017framework}, and BHT~\cite{muniraj2017framework}. CUSUM and SPRT are derived from a probabilistic perspective and exhibit good performance in detecting single attacks. However, when applied to UAVs, where long-term performance is crucial, their effectiveness is significantly reduced and we need to tune the threshold parameters for each case to achieve the best performance. As depicted in Fig.~\ref{fig:CUSUM_3_plots}, the CUSUM detector continues to generate false alarm modes even after the attack has ceased. This occurs because constant attacks introduce a bias in the estimation with a constant offset.
Consequently, when the attack subsides, the residue's square norm resembles that observed during the attack, leading to false alarms. On the other hand, the BHT assumes that in the absence of attacks, the residue will predominantly lie within an ellipsoid. This assumption is rather strong, as in reality, the shape depends on the system model and noise, and may not necessarily conform to an ellipsoid. Overall, the BHT can be regarded as a special case of SVM, where the hyperplane is represented by a limited ellipsoid. 

In order to ensure fair comparisons, we replace the original inputs of existing learning-based methods with the residue sequence produced by our residue generator. This substitution is necessary because using raw sensor data as input, which can have arbitrary values and ranges, would not result in expected performance under our complex UAV models. Furthermore, each method is trained accurately to convergence and its parameters are adjusted as much as possible to improve performance. SVM~\cite{shafique2021detecting}\cite{khoei2022comparative} based on kernel techniques has nonlinear high-dimensional classification capabilities, and CNN~\cite{wu2023highly} and LSTM~\cite{wu2023highly}\cite{wang2020intelligent} have the ability to extract local features and short-term information. 
\begin{figure}[h]
    \centering
    \hfill 
    \begin{subfigure}[b]{0.49\columnwidth}
        \includegraphics[width=\linewidth]{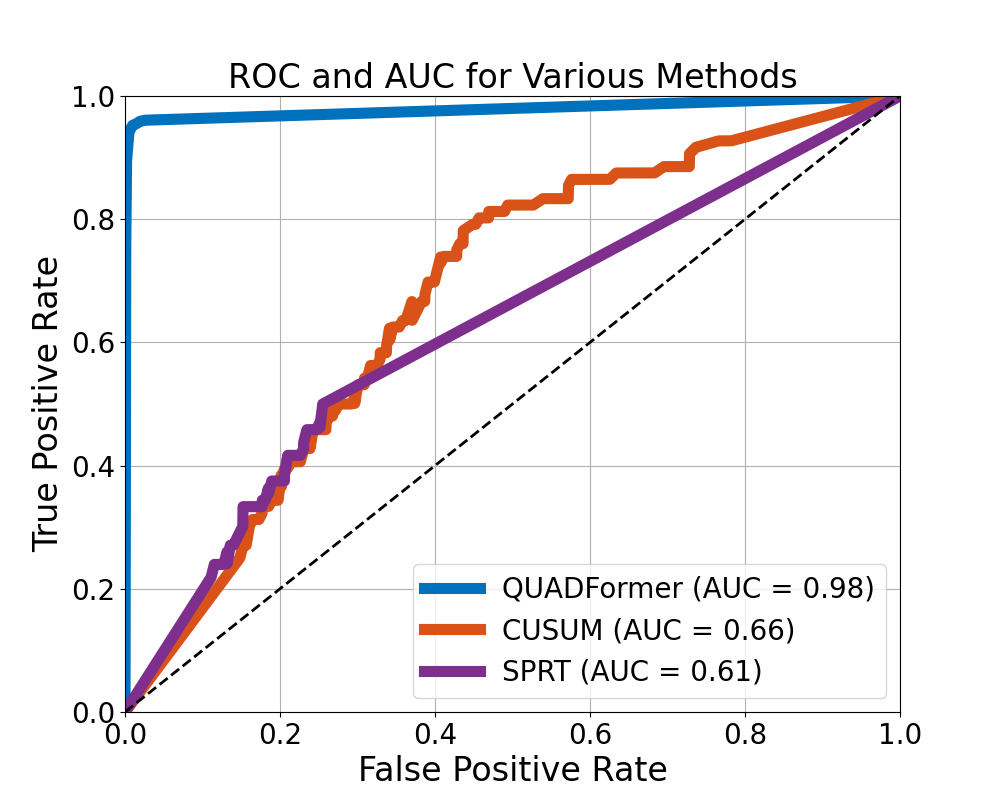}
        \caption{Traditional methods.}
    \end{subfigure}
    \hfill
    \begin{subfigure}[b]{0.49\columnwidth}
        \includegraphics[width=\linewidth]{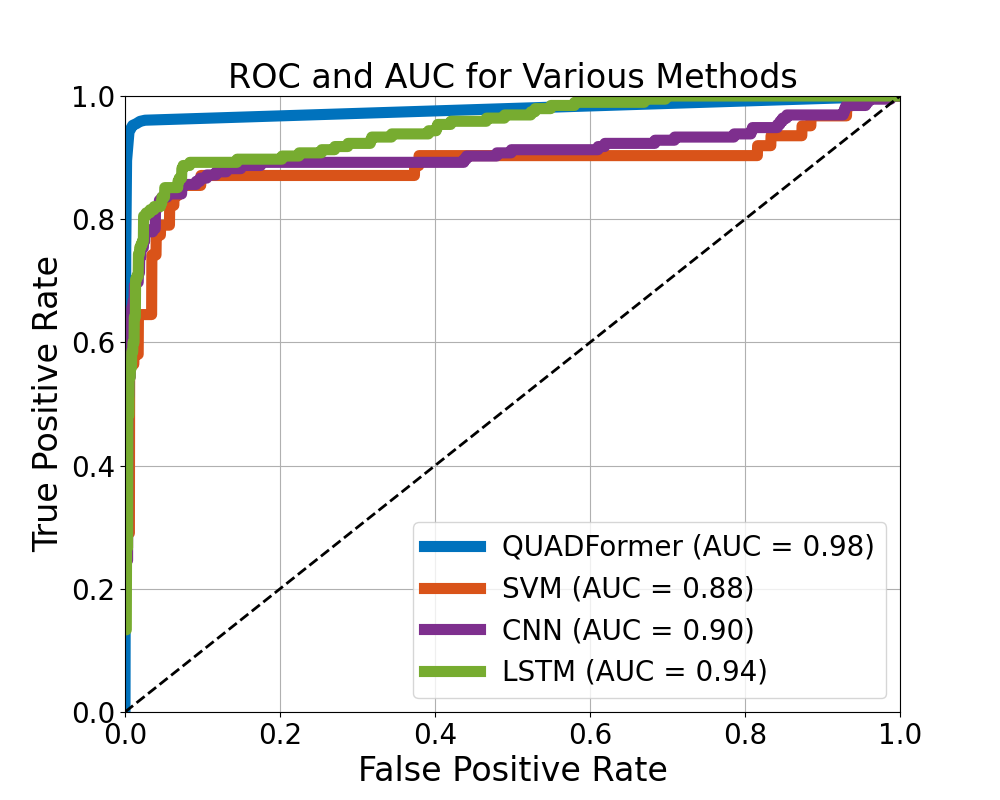}
        \caption{Learning methods.}
    \end{subfigure}
    \caption{ROC/AUC comparison results for Model I.}
    \label{fig:roc}
\end{figure}

\begin{figure}[h]
    \centering
    \hfill 
    \begin{subfigure}[b]{0.49\columnwidth}
        \includegraphics[width=\linewidth]{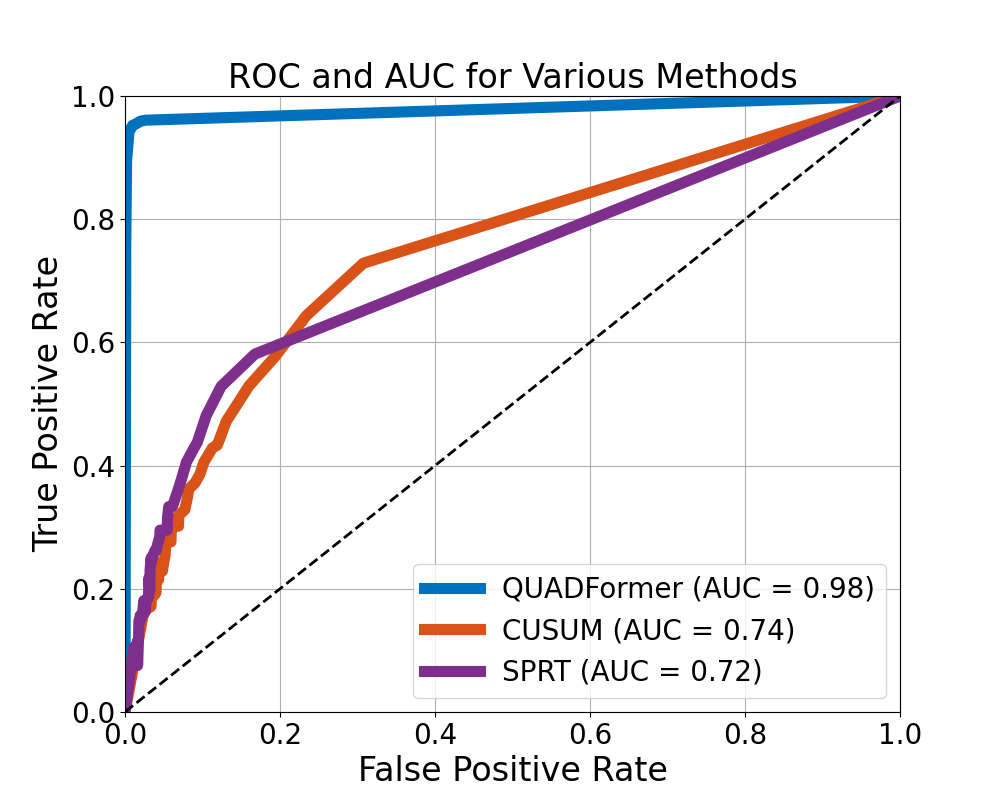}
        \caption{Traditional methods.}
    \end{subfigure}
    \hfill
    \begin{subfigure}[b]{0.49\columnwidth}
        \includegraphics[width=\linewidth]{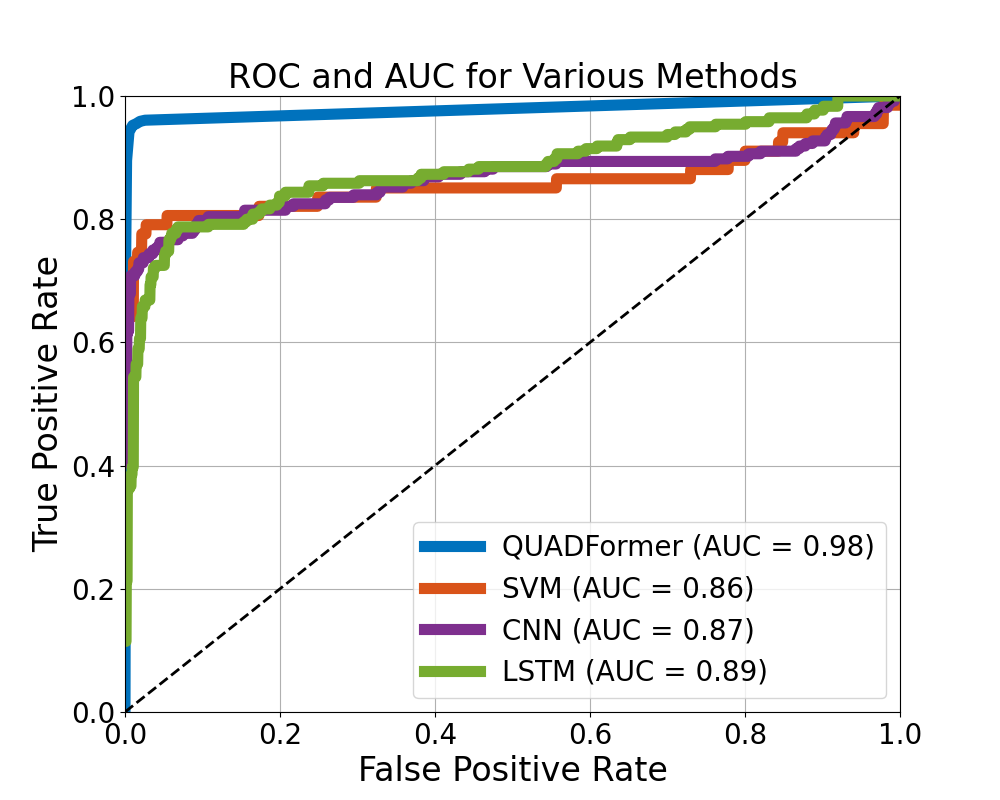}
        \caption{Learning methods.}
    \end{subfigure}
    \caption{ROC/AUC comparison results for Model II.}
    \label{fig:roc2}
\end{figure}
Since attacks on UAVs are sparse, several learning-based methods have similar performance to our method in terms of \textit{Precision}. However, unlike window-based CNN or LSTM which process information step by step, our method is more effective in identifying complex patterns across time steps, by utilizing a special attention calculation method. At the same time, due to the parallel processing and dynamic weight capabilities of the transformer itself, our method has significant advantages in the two indicators of \textit{Recall} and \textit{F1-score}, especially when the attack amplitude is small and stealthy. Finally, we calculate and visualize the Receiver Operating Characteristic (ROC) curve and Area Under the Curve (AUC) of all the related methods to show the efficiency and effectiveness of our method, and two representative results are shown in Fig.~\ref{fig:roc} - Fig.~\ref{fig:roc2}.

% The first table
\begin{table*}[h]
\centering
\caption{Comparison Results for Model I.}
\label{tab:comprehensive_analysis1}

\bgroup
\renewcommand{\arraystretch}{1.5} 

\begin{tabularx}{\textwidth}{|c|c|>{\centering\arraybackslash}X|>{\centering\arraybackslash}X|>{\centering\arraybackslash}X|>{\centering\arraybackslash}X|>{\centering\arraybackslash}X|>{\centering\arraybackslash}X|>{\centering\arraybackslash}X|>{\centering\arraybackslash}X|>{\centering\arraybackslash}X|>{\centering\arraybackslash}X|>{\centering\arraybackslash}X|>{\centering\arraybackslash}X|}
\hline
\multirow{3}{*}{System Model} & UAV Model & \multicolumn{12}{c|}{Model I} \\
\cline{2-14}
 & Noise Model & \multicolumn{6}{c|}{Exponential Noise} & \multicolumn{6}{c|}{Laplacian Noise} \\
\cline{2-14}
 & Attack Model & \multicolumn{3}{c|}{Attack I} & \multicolumn{3}{c|}{Attack II} & \multicolumn{3}{c|}{Attack I} & \multicolumn{3}{c|}{Attack II} \\
\hline
Performance & Metric & P & R & F1 & P & R & F1 & P & R & F1 & P & R & F1 \\
\hline
\multirow{3}{*}{Traditional Method} & CUSUM\cite{yoon2019towards}\cite{liu2019secure} & 0.69 & 0.64 & 0.66 & 0.71 & 0.59 & 0.64 & 0.70 & 0.62 & 0.66 & 0.71 & 0.58 & 0.64 \\
\cline{2-14}
 & SPRT\cite{muniraj2017framework} & 0.69 & 0.67 & 0.69 & 0.68 & 0.68 & 0.67 & 0.69 & 0.67 & 0.68 & 0.68 & 0.67 & 0.67 \\
\cline{2-14}
 & BHT\cite{muniraj2017framework} & 0.62 & 0.65 & 0.64 & 0.68 & 0.60 & 0.65 & 0.66 & 0.59 & 0.62 & 0.70 & 0.60 & 0.65 \\
\hline
\multirow{3}{*}{\parbox{2cm}{\centering Learning-based\\ Method}} & SVM\cite{shafique2021detecting}\cite{khoei2022comparative} & 0.88 & 0.82 & 0.85 & 0.75 & 0.69 & 0.72 & 0.88 & 0.82 & 0.85 & 0.75 & 0.69 & 0.72 \\
\cline{2-14}
 & CNN\cite{wu2023highly} & 0.89 & 0.81 & 0.85 & 0.90 & 0.78 & 0.84 & 0.90 & 0.84 & 0.87 & 0.90 & 0.78 & 0.84 \\
\cline{2-14}
 & LSTM\cite{wu2023highly}\cite{wang2020intelligent} & 0.88 & 0.78 & 0.83 & 0.78 & 0.86 & 0.82 & 0.86 & 0.84 & 0.85 & 0.81 & 0.85 & 0.83 \\
\hline
\textbf{Ours} & \textbf{QUADFormer} & \textbf{0.92} & \textbf{0.93} & \textbf{0.93} & \textbf{0.89} & \textbf{0.95} & \textbf{0.92} & \textbf{0.92} & \textbf{0.94} & \textbf{0.93} & \textbf{0.89} & \textbf{0.95} & \textbf{0.92} \\
\hline
\end{tabularx}

\egroup 
\end{table*}

% The second table
\begin{table*}[h]
\centering
\caption{Comparison Results for Model II.}
\label{tab:comprehensive_analysis2}

\bgroup
\renewcommand{\arraystretch}{1.5} 

\begin{tabularx}{\textwidth}{|c|c|>{\centering\arraybackslash}X|>{\centering\arraybackslash}X|>{\centering\arraybackslash}X|>{\centering\arraybackslash}X|>{\centering\arraybackslash}X|>{\centering\arraybackslash}X|>{\centering\arraybackslash}X|>{\centering\arraybackslash}X|>{\centering\arraybackslash}X|>{\centering\arraybackslash}X|>{\centering\arraybackslash}X|>{\centering\arraybackslash}X|}
\hline
\multirow{3}{*}{System Model} & UAV Model & \multicolumn{12}{c|}{Model II} \\
\cline{2-14}
 & Noise Model & \multicolumn{6}{c|}{Exponential Noise} & \multicolumn{6}{c|}{Laplacian Noise} \\
\cline{2-14}
 & Attack Model & \multicolumn{3}{c|}{Attack I} & \multicolumn{3}{c|}{Attack II} & \multicolumn{3}{c|}{Attack I} & \multicolumn{3}{c|}{Attack II} \\
\hline
Performance & Metric & P & R & F1 & P & R & F1 & P & R & F1 & P & R & F1 \\
\hline
\multirow{3}{*}{Traditional Method} & CUSUM\cite{yoon2019towards}\cite{liu2019secure} & 0.75 & 0.76 & 0.76 & 0.73 & 0.70 & 0.72 & 0.73 & 0.69 & 0.71 & 0.72 & 0.69 & 0.71 \\
\cline{2-14}
 & SPRT\cite{muniraj2017framework} & 0.75 & 0.77 & 0.76 & 0.74 & 0.77 & 0.74 & 0.74 & 0.73 & 0.73 & 0.73 & 0.70 & 0.71 \\
\cline{2-14}
 & BHT\cite{muniraj2017framework} & 0.70 & 0.72 & 0.72 & 0.75 & 0.71 & 0.73 & 0.75 & 0.70 & 0.72 & 0.69 & 0.72 & 0.71 \\
\hline
\multirow{3}{*}{\parbox{2cm}{\centering Learning-based\\ Method}} & SVM\cite{shafique2021detecting}\cite{khoei2022comparative} & 0.94 & 0.76 & 0.84 & 0.94 & 0.61 & 0.74 & 0.94 & 0.75 & 0.83 & 0.93 & 0.61 & 0.74 \\
\cline{2-14}
 & CNN\cite{wu2023highly} & 0.93 & 0.73 & 0.82 & 0.94 & 0.70 & 0.80 & 0.96 & 0.67 & 0.79 & 0.85 & 0.72 & 0.78 \\
\cline{2-14}
 & LSTM\cite{wu2023highly}\cite{wang2020intelligent} & 0.89 & 0.80 & 0.84 & 0.90 & 0.75 & 0.82 & 0.81 & 0.83 & 0.82 & 0.89 & 0.74 & 0.81 \\
\hline
\textbf{Ours} & \textbf{QUADFormer} & \textbf{0.92} & \textbf{0.97} & \textbf{0.94} & \textbf{0.89} & \textbf{0.94} & \textbf{0.91} & \textbf{0.91} & \textbf{0.96} & \textbf{0.94} & \textbf{0.89} & \textbf{0.97} & \textbf{0.93} \\
\hline
\end{tabularx}

\egroup 
\end{table*}

\section{Conclusion}

In this paper, we have presented a novel cyber attack detection framework specifically designed for quadrotor UAVs, integrating both residue-based and learning-based methods. This integration has effectively addressed the shortcomings previously observed in standalone methods. We have successfully leveraged this framework to extract and learn the statistical characteristics of the UAV's highly nonlinear dynamics and non-Gaussian noises. As a result, our approach has achieved exceptional performance in detecting sensor attacks compared with existing methods. In the future, We will conduct experiments on more types of cyber attacks, such as random attacks, reply attacks, etc., and we will also explore different ways to generate better residue sequences for more complex systems.

\section*{Acknowledgment}
The authors appreciate the kind help from Jianping Zhou (SJTU). The authors also appreciate the open-source code provided in~\cite{tuli2022tranad}~\cite{xu2022anomaly} and the course materials of HKUST ELEC 5660.

\bibliographystyle{IEEEtran}
\bibliography{main}

\begin{thebibliography}{10}
\providecommand{\url}[1]{#1}
\csname url@samestyle\endcsname
\providecommand{\newblock}{\relax}
\providecommand{\bibinfo}[2]{#2}
\providecommand{\BIBentrySTDinterwordspacing}{\spaceskip=0pt\relax}
\providecommand{\BIBentryALTinterwordstretchfactor}{4}
\providecommand{\BIBentryALTinterwordspacing}{\spaceskip=\fontdimen2\font plus
\BIBentryALTinterwordstretchfactor\fontdimen3\font minus \fontdimen4\font\relax}
\providecommand{\BIBforeignlanguage}[2]{{%
\expandafter\ifx\csname l@#1\endcsname\relax
\typeout{** WARNING: IEEEtran.bst: No hyphenation pattern has been}%
\typeout{** loaded for the language `#1'. Using the pattern for}%
\typeout{** the default language instead.}%
\else
\language=\csname l@#1\endcsname
\fi
#2}}
\providecommand{\BIBdecl}{\relax}
\BIBdecl

\bibitem{alur2015principles}
R.~Alur, \emph{Principles of Cyber-physical Systems}.\hskip 1em plus 0.5em minus 0.4em\relax MIT press, 2015.

\bibitem{li2019coverage}
T.~Li, C.~Wang, and C.~W. de~Silva, ``Coverage sampling planner for {UAV}-enabled environmental exploration and field mapping,'' in \emph{IEEE/RSJ International Conference on Intelligent Robots and Systems}, 2019, pp. 2509--2516.

\bibitem{ngo2022uav}
E.~Ngo, J.~Ramirez, M.~Medina-Soto, S.~Dirksen, E.~D. Victoriano, and S.~Bhandari, ``{UAV} platforms for autonomous navigation in {GPS}-denied environments for search and rescue missions,'' in \emph{IEEE International Conference on Unmanned Aircraft Systems}, 2022, pp. 1481--1488.

\bibitem{wang2022quadrotor}
P.~Wang, C.~Wang, J.~Wang, and M.~Q.-H. Meng, ``Quadrotor autonomous landing on moving platform,'' \emph{Procedia Computer Science}, vol. 209, pp. 40--49, 2022.

\bibitem{restivo2023gps}
R.~D. Restivo, L.~C. Dodson, J.~Wang, W.~Tan, Y.~Liu, H.~Wang, and H.~Song, ``{GPS} spoofing on {UAV}: A survey,'' in \emph{IEEE INFOCOM 2023-IEEE Conference on Computer Communications Workshops}, 2023, pp. 1--6.

\bibitem{adil2023uav}
M.~Adil, H.~Song, S.~Mastorakis, H.~Abulkasim, A.~Farouk, and Z.~Jin, ``{UAV}-assisted {IoT} applications, cybersecurity threats, {AI}-enabled solutions, open challenges with future research directions,'' \emph{IEEE Transactions on Intelligent Vehicles}, 2023.

\bibitem{shepard2012evaluation}
D.~P. Shepard, J.~A. Bhatti, T.~E. Humphreys, and A.~A. Fansler, ``Evaluation of smart grid and civilian {UAV} vulnerability to {GPS} spoofing attacks,'' in \emph{Proceedings of the 25th International Technical Meeting of The Satellite Division of the Institute of Navigation}, 2012, pp. 3591--3605.

\bibitem{kerns2014unmanned}
A.~J. Kerns, D.~P. Shepard, J.~A. Bhatti, and T.~E. Humphreys, ``Unmanned aircraft capture and control via {GPS} spoofing,'' \emph{Journal of Field Robotics}, vol.~31, no.~4, pp. 617--636, 2014.

\bibitem{hooper2016securing}
M.~Hooper, Y.~Tian, R.~Zhou, B.~Cao, A.~P. Lauf, L.~Watkins, W.~H. Robinson, and W.~Alexis, ``Securing commercial {WiFi}-based {UAVs} from common security attacks,'' in \emph{IEEE Military Communications Conference}, 2016, pp. 1213--1218.

\bibitem{dey2018security}
V.~Dey, V.~Pudi, A.~Chattopadhyay, and Y.~Elovici, ``Security vulnerabilities of unmanned aerial vehicles and countermeasures: An experimental study,'' in \emph{31st IEEE International Conference on VLSI Design and 17th International Conference on Embedded Systems}, 2018, pp. 398--403.

\bibitem{gregory2013cyber}
M.~A. Gregory, D.~Glance, M.~A. Gregory, and D.~Glance, ``Cyber crime, cyber security and cyber warfare,'' \emph{Security and the Networked Society}, pp. 51--95, 2013.

\bibitem{hassija2021fast}
V.~Hassija, V.~Chamola, A.~Agrawal, A.~Goyal, N.~C. Luong, D.~Niyato, F.~R. Yu, and M.~Guizani, ``Fast, reliable, and secure drone communication: A comprehensive survey,'' \emph{IEEE Communications Surveys \& Tutorials}, vol.~23, no.~4, pp. 2802--2832, 2021.

\bibitem{griffioen2019tutorial}
P.~Griffioen, S.~Weerakkody, B.~Sinopoli, O.~Ozel, and Y.~Mo, ``A tutorial on detecting security attacks on cyber-physical systems,'' in \emph{18th European Control Conference}, 2019, pp. 979--984.

\bibitem{chen2019container}
J.~Chen, Z.~Feng, J.-Y. Wen, B.~Liu, and L.~Sha, ``A container-based {DoS} attack-resilient control framework for real-time {UAV} systems,'' in \emph{IEEE Design, Automation \& Test in Europe Conference \& Exhibition}, 2019, pp. 1222--1227.

\bibitem{chen2017false}
W.~Chen, Z.~Duan, and Y.~Dong, ``False data injection on {EKF}-based navigation control,'' in \emph{IEEE International Conference on Unmanned Aircraft Systems}, 2017, pp. 1608--1617.

\bibitem{porter2020detecting}
M.~Porter, P.~Hespanhol, A.~Aswani, M.~Johnson-Roberson, and R.~Vasudevan, ``Detecting generalized replay attacks via time-varying dynamic watermarking,'' \emph{IEEE Transactions on Automatic Control}, vol.~66, no.~8, pp. 3502--3517, 2020.

\bibitem{fang2020optimal}
C.~Fang, Y.~Qi, P.~Cheng, and W.~X. Zheng, ``Optimal periodic watermarking schedule for replay attack detection in cyber-physical systems,'' \emph{Automatica}, vol. 112, p. 108698, 2020.

\bibitem{murguia2016cusum}
C.~Murguia and J.~Ruths, ``{CUSUM} and chi-squared attack detection of compromised sensors,'' in \emph{IEEE Conference on Control Applications}, 2016, pp. 474--480.

\bibitem{murguia2016characterization}
------, ``Characterization of a {CUSUM} model-based sensor attack detector,'' in \emph{IEEE 55th Conference on Decision and Control}, 2016, pp. 1303--1309.

\bibitem{yoon2019towards}
H.-J. Yoon, W.~Wan, H.~Kim, N.~Hovakimyan, L.~Sha, and P.~G. Voulgaris, ``Towards resilient {UAV}: Escape time in {GPS} denied environment with sensor drift,'' \emph{IFAC-PapersOnLine}, vol.~52, no.~12, pp. 423--428, 2019.

\bibitem{liu2019secure}
Q.~Liu, Y.~Mo, X.~Mo, C.~Lv, E.~Mihankhah, and D.~Wang, ``Secure pose estimation for autonomous vehicles under cyber attacks,'' in \emph{IEEE Intelligent Vehicles Symposium}, 2019, pp. 1583--1588.

\bibitem{muniraj2017framework}
D.~Muniraj and M.~Farhood, ``A framework for detection of sensor attacks on small unmanned aircraft systems,'' in \emph{IEEE International Conference on Unmanned Aircraft Systems}, 2017, pp. 1189--1198.

\bibitem{wan2020safety}
W.~Wan, H.~Kim, N.~Hovakimyan, L.~Sha, and P.~G. Voulgaris, ``A safety constrained control framework for {UAV}s in {GPS} denied environment,'' in \emph{59th IEEE Conference on Decision and Control}, 2020, pp. 214--219.

\bibitem{farivar2019artificial}
F.~Farivar, M.~S. Haghighi, A.~Jolfaei, and M.~Alazab, ``Artificial intelligence for detection, estimation, and compensation of malicious attacks in nonlinear cyber-physical systems and industrial {IoT},'' \emph{IEEE Transactions on Industrial Informatics}, vol.~16, no.~4, pp. 2716--2725, 2019.

\bibitem{shen2020timeseries}
L.~Shen, Z.~Li, and J.~Kwok, ``Timeseries anomaly detection using temporal hierarchical one-class network,'' \emph{Advances in Neural Information Processing Systems}, vol.~33, pp. 13\,016--13\,026, 2020.

\bibitem{shafique2021detecting}
A.~Shafique, A.~Mehmood, and M.~Elhadef, ``Detecting signal spoofing attack in {UAVs} using machine learning models,'' \emph{IEEE Access}, vol.~9, pp. 93\,803--93\,815, 2021.

\bibitem{khoei2022comparative}
T.~T. Khoei, A.~Gasimova, M.~A. Ahajjam, K.~Al~Shamaileh, V.~Devabhaktuni, and N.~Kaabouch, ``A comparative analysis of supervised and unsupervised models for detecting {GPS} spoofing attack on {UAVs},'' in \emph{IEEE International Conference on Electro Information Technology}, 2022, pp. 279--284.

\bibitem{wu2023highly}
S.~Wu, Y.~Li, Z.~Wang, Z.~Tan, and Q.~Pan, ``A highly interpretable framework for generic low-cost {UAV} attack detection,'' \emph{IEEE Sensors Journal}, vol.~23, no.~7, pp. 7288--7300, 2023.

\bibitem{wang2020intelligent}
S.~Wang, J.~Wang, C.~Su, and X.~Ma, ``Intelligent detection algorithm against {UAVs}' {GPS} spoofing attack,'' in \emph{IEEE 26th International Conference on Parallel and Distributed Systems}, 2020, pp. 382--389.

\bibitem{xiao2022cyber}
J.~Xiao and M.~Feroskhan, ``Cyber attack detection and isolation for a quadrotor {UAV} with modified sliding innovation sequences,'' \emph{IEEE Transactions on Vehicular Technology}, vol.~71, no.~7, pp. 7202--7214, 2022.

\bibitem{yang2023lasso}
N.~Yang, Z.~Wang, T.~Yang, Y.~Li, and L.~Shi, ``{LASSO}-based detection and identification of actuator integrity attacks in remote control systems,'' \emph{IEEE Transactions on Control Systems Technology}, 2023.

\bibitem{li2021multivariate}
Z.~Li, Y.~Zhao, J.~Han, Y.~Su, R.~Jiao, X.~Wen, and D.~Pei, ``Multivariate time series anomaly detection and interpretation using hierarchical inter-metric and temporal embedding,'' in \emph{Proceedings of the 27th ACM SIGKDD Conference on Knowledge Discovery \& Data Mining}, 2021, pp. 3220--3230.

\bibitem{tuli2022tranad}
S.~Tuli, G.~Casale, and N.~R. Jennings, ``Tranad: Deep transformer networks for anomaly detection in multivariate time series data,'' \emph{Proceedings of VLDB}, vol.~15, no.~6, pp. 1201--1214, 2022.

\bibitem{xu2022anomaly}
J.~Xu, H.~Wu, J.~Wang, and M.~Long, ``Anomaly transformer: Time series anomaly detection with association discrepancy,'' in \emph{International Conference on Learning Representations}, 2022.

\bibitem{vaswani2017attention}
A.~Vaswani, N.~Shazeer, N.~Parmar, J.~Uszkoreit, L.~Jones, A.~N. Gomez, {\L}.~Kaiser, and I.~Polosukhin, ``Attention is all you need,'' \emph{Advances in Neural Information Processing Systems}, vol.~30, 2017.

\bibitem{abbaspour2017neural}
A.~Abbaspour, P.~Aboutalebi, K.~K. Yen, and A.~Sargolzaei, ``Neural adaptive observer-based sensor and actuator fault detection in nonlinear systems: Application in {UAV},'' \emph{ISA Transactions}, vol.~67, pp. 317--329, 2017.

\bibitem{xiao2017user}
L.~Xiao, C.~Xie, M.~Min, and W.~Zhuang, ``User-centric view of unmanned aerial vehicle transmission against smart attacks,'' \emph{IEEE Transactions on Vehicular Technology}, vol.~67, no.~4, pp. 3420--3430, 2017.

\bibitem{ma2023detecting}
Y.~Ma, M.~N. Al~Islam, J.~Cleland-Huang, and N.~V. Chawla, ``Detecting anomalies in small unmanned aerial systems via graphical normalizing flows,'' \emph{IEEE Intelligent Systems}, 2023.

\bibitem{abbaspour2018neural}
A.~Abbaspour, K.~K. Yen, P.~Forouzannezhad, and A.~Sargolzaei, ``A neural adaptive approach for active fault-tolerant control design in {UAV},'' \emph{IEEE Transactions on Systems, Man, and Cybernetics: Systems}, vol.~50, no.~9, pp. 3401--3411, 2018.

\bibitem{hickling2023robust}
T.~Hickling, N.~Aouf, and P.~Spencer, ``Robust adversarial attacks detection based on explainable deep reinforcement learning for {UAV} guidance and planning,'' \emph{IEEE Transactions on Intelligent Vehicles}, 2023.

\bibitem{menegaz2015systematization}
H.~M. Menegaz, J.~Y. Ishihara, G.~A. Borges, and A.~N. Vargas, ``A systematization of the unscented {Kalman} filter theory,'' \emph{IEEE Transactions on Automatic Control}, vol.~60, no.~10, pp. 2583--2598, 2015.

\bibitem{gustafsson2010particle}
F.~Gustafsson, ``Particle filter theory and practice with positioning applications,'' \emph{IEEE Aerospace and Electronic Systems Magazine}, vol.~25, no.~7, pp. 53--82, 2010.

\bibitem{reif1999stochastic}
K.~Reif, S.~Gunther, E.~Yaz, and R.~Unbehauen, ``Stochastic stability of the discrete-time extended kalman filter,'' \emph{IEEE Transactions on Automatic control}, vol.~44, no.~4, pp. 714--728, 1999.

\bibitem{chen2021crossvit}
C.-F.~R. Chen, Q.~Fan, and R.~Panda, ``Crossvit: Cross-attention multi-scale vision transformer for image classification,'' in \emph{Proceedings of the IEEE/CVF International Conference on Computer Vision}, 2021, pp. 357--366.

\bibitem{mourikis2007multi}
A.~I. Mourikis and S.~I. Roumeliotis, ``A multi-state constraint kalman filter for vision-aided inertial navigation,'' in \emph{IEEE International Conference on Robotics and Automation}, 2007, pp. 3565--3572.

\bibitem{carrillo2014state}
L.~R.~G. Carrillo, W.~J. Russell, J.~P. Hespanha, and G.~E. Collins, ``State estimation of multiagent systems under impulsive noise and disturbances,'' \emph{IEEE Transactions on Control Systems Technology}, vol.~23, no.~1, pp. 13--26, 2014.

\end{thebibliography}

\end{document}